\documentclass[11pt]{article}

\usepackage{a4}
\usepackage{a4wide}
\usepackage[dvips]{graphicx}
\usepackage{xspace}
\usepackage{avm}
\usepackage{examples}
\usepackage{amsmath}
\usepackage{amssymb}

\renewcommand{\paragraph}[1]{\vspace{2mm}\noindent\textbf{#1}}

\newcommand{\feat}{\sc}                         
\newcommand{\fval}{\it}                         
\newcommand{\srt}{\it}                          
\newcommand{\pref}[1]{(\ref{#1})}               
\newcommand{\qit}[1]{\textsl{``#1''}}           
\newcommand{\sys}[1]{\texttt{#1}}               
\newcommand{\sql}[1]{{\small \texttt{#1}}}      
\newcommand{\newt}[1]{\textit{#1}}              

\newcommand{\nldb}{\textsc{nldb}\xspace}
\newcommand{\nltdb}{\textsc{nltdb}\xspace}
\newcommand{\hpsg}{\textsc{hpsg}\xspace}

\newcommand{\topl}{\textsc{top}\xspace}
\newcommand{\tsql}{\textsc{tsql}{\footnotesize 2}\xspace}
\newcommand{\sqll}{\textsc{sql}\xspace}
\newcommand{\ale}{\textsc{ale}\xspace}
\newcommand{\timedb}{\textsc{timedb}\xspace}

\newcommand{\dbms}{\textsc{dbms}\xspace}
\newcommand{\ils}{\textsc{il}$_s$\xspace}
\newcommand{\ptq}{\textsc{ptq}\xspace}
\newcommand{\dcg}{\textsc{dcg}\xspace}

\newcommand{\bad}{\sqz{*}}                           

\newcommand{\subper}{\ensuremath{\sqsubseteq}}       
\newcommand{\propsubper}{\ensuremath{\sqsubset}}     
\newcommand{\union}{\cup}                            
\newcommand{\intersect}{\cap}                        
\newcommand{\defeq}{\equiv}                          
\newcommand{\denot}[2]{\|#2\|^{#1}}                  
\newcommand{\tup}[1]{\langle#1\rangle}               

\newcommand{\partop}{\ensuremath{\mathit{Part}}\xspace}
\newcommand{\pres}{\ensuremath{\mathit{Pres}}\xspace}
\newcommand{\past}{\ensuremath{\mathit{Past}}\xspace}
\newcommand{\culm}{\ensuremath{\mathit{Culm}}\xspace}
\newcommand{\perf}{\ensuremath{\mathit{Perf}}\xspace}

\newcommand{\at}{\ensuremath{\mathit{At}}\xspace}
\newcommand{\ntense}{\ensuremath{\mathit{Ntense}}\xspace}
\newcommand{\before}{\ensuremath{\mathit{Before}}\xspace}
\newcommand{\after}{\ensuremath{\mathit{After}}\xspace}
\newcommand{\lbegin}{\ensuremath{\mathit{Begin}}\xspace}
\newcommand{\lend}{\ensuremath{\mathit{End}}\xspace}
\newcommand{\fills}{\ensuremath{\mathit{Fills}}\xspace}
\newcommand{\for}{\ensuremath{\mathit{For}}\xspace}

\newcommand{\literal}{\ensuremath{\mathit{LITERAL}}}

\newcommand{\pts}{\ensuremath{\mathit{PTS}}\xspace}
\newcommand{\periods}{\ensuremath{\mathit{PERIODS}}\xspace}

\newcommand{\instants}{\ensuremath{\mathit{INSTANTS}}\xspace}
\newcommand{\cons}{\ensuremath{\mathit{CONS}}\xspace}
\newcommand{\vars}{\ensuremath{\mathit{VARS}}\xspace}
\newcommand{\terms}{\ensuremath{\mathit{TERMS}}\xspace}
\newcommand{\pfuns}{\ensuremath{\mathit{PFUNS}}\xspace}
\newcommand{\parts}{\ensuremath{\mathit{PARTS}}\xspace}
\newcommand{\cparts}{\ensuremath{\mathit{CPARTS}}\xspace}
\newcommand{\gparts}{\ensuremath{\mathit{GPARTS}}\xspace}
\newcommand{\aforms}{\ensuremath{\mathit{AFORMS}}\xspace}
\newcommand{\forms}{\ensuremath{\mathit{FORMS}}\xspace}
\newcommand{\ynforms}{\ensuremath{\mathit{YNFORMS}}\xspace}
\newcommand{\whformsone}{\ensuremath{\mathit{WHFORMS1}}\xspace}
\newcommand{\whformstwo}{\ensuremath{\mathit{WHFORMS2}}\xspace}
\newcommand{\whforms}{\ensuremath{\mathit{WHFORMS}}\xspace}
\newcommand{\objs}{\ensuremath{\mathit{OBJS}}\xspace}
\newcommand{\fcons}{\ensuremath{\mathit{f_{cons}}}\xspace}
\newcommand{\fpfuns}{\ensuremath{\mathit{f_{pfuns}}}\xspace}
\newcommand{\fculms}{\ensuremath{\mathit{f_{culms}}}\xspace}
\newcommand{\fgparts}{\ensuremath{\mathit{f_{gparts}}}\xspace}
\newcommand{\fcparts}{\ensuremath{\mathit{f_{cparts}}}\xspace}
\newcommand{\pow}{\ensuremath{\mathit{pow}}\xspace}
\newcommand{\mxlpers}{\ensuremath{\mathit{mxlpers}}\xspace}

\newcommand{\hconsp}{\ensuremath{\mathit{h'_{cons}}}\xspace}

\newcommand{\hpfunsp}{\ensuremath{\mathit{h'_{pfuns}}}\xspace}

\newcommand{\dbtable}[5]{{\small\begin{tabular}[t]{#2}
      \hline                            
      \multicolumn{#1}{|l|}{#3} \\      
      \hline \hline                     
      #4 \\
      \hline
      #5 \\
      \hline 
   \end{tabular}}}

\newcommand{\dbtableb}[2]{{\small\begin{tabular}[t]{#1}
      \hline                            
      #2 \\
      \hline 
   \end{tabular}}}

\newcommand{\adbtable}[5]        
{\vspace{-3mm}
 \begin{center}
 \dbtable{#1}{#2}{#3}{#4}{#5} 
 \end{center}}

\newcommand{\select}[1]                
{\sql{\hspace*{-2mm}\begin{tabular}[t]{l}
#1
\end{tabular}}}

\avmfont{\footnotesize\feat}
\avmvalfont{\footnotesize\fval}
\avmsortfont{\scriptsize\srt}

\newbox\avmboxa
\newbox\avmboxb
\newbox\avmboxc 

\newcommand{\gnote}[1]{\ensuremath{\langle #1\rangle}}

\title{Time, Tense and Aspect in Natural Language Database Interfaces\thanks{
    This paper reports on work that was
    carried out while the first author was in the Department of Artificial
    Intelligence, University of Edinburgh, supported by the Greek
    State Scholarships Foundation.}}
\author{
    Ion Androutsopoulos$^1$
    \and 
    Graeme Ritchie$^2$
    \and 
    Peter Thanisch$^3$} 
\date{$^1$Language Technology Group, Microsoft Research Institute, \\
      Macquarie University, Sydney NSW 2109, Australia \\
      e-mail: \texttt{ion@mri.mq.edu.au} 
      \vspace{3mm}\\
      $^2$Department of Artificial Intelligence, 
      University of Edinburgh \\
      80 South Bridge, Edinburgh EH1 1HN, Scotland, U.K. \\
      e-mail: \texttt{G.D.Ritchie@ed.ac.uk}
      \vspace{3mm}\\
      $^3$Department of Computer Science, University of Edinburgh \\
      King's Buildings, Mayfield Road, Edinburgh EH9 3JZ, Scotland, U.K. \\
      e-mail: \texttt{pt@dcs.ed.ac.uk}
     } 

\begin{document}
\maketitle


\begin{abstract}

\textsl{\small
Most existing natural language database interfaces (\nldb{s}) were
designed to be used with database systems that provide
very limited facilities for manipulating time-dependent
data, and they do not support adequately temporal linguistic
mechanisms (verb tenses, temporal adverbials, temporal subordinate
clauses, etc.). The database community is becoming increasingly
interested in temporal database systems, that are intended to store
and manipulate in a principled manner information not only about the
present, but also about the past and future.  When interfacing to
temporal databases, supporting temporal linguistic mechanisms becomes
crucial.}

\textsl{\small
We present a framework for constructing natural language interfaces for
temporal databases (\nltdb{s}), that draws on research in tense
and aspect theories, temporal logics, and temporal databases. The
framework consists of a temporal intermediate representation
language, called \topl, an \hpsg grammar that maps a wide range of
questions involving temporal mechanisms to appropriate \topl
expressions, and a provably correct method for translating from \topl to
\tsql, \tsql being a recently proposed temporal extension of the \sqll database
language. This framework was employed to implement a prototype \nltdb.
}
\end{abstract}


\section{Introduction} \label{introduction}

Since the 1960s, natural language database interfaces (\nldb{s}) 
have been the subject of much attention in the natural language 
processing community  \cite{Perrault},
\cite{Copestake}, \cite{Androutsopoulos1995}.
\nldb{s} allow users to access information stored in databases by formulating
requests in natural language. Most existing \nldb{s} were designed to
interface to database systems that provide very limited facilities for
manipulating time-dependent data. Consequently, most \nldb{s} also
provide very limited temporal support. In particular, they are
intended to answer questions that 
make only a superficial use of time, in the form of timestamp
values that are just treated as a form of numerical data by the
database language, and do
not adequately support temporal linguistic mechanisms (verb tenses and
aspects, temporal adverbials, temporal subordinate clauses, etc.):
users are allowed to use very few (if any) of these mechanisms, and
their semantics are typically over-simplified or even ignored.

The database community is becoming increasingly interested in
\emph{temporal} database systems. These are intended to store and
manipulate in a principled manner information not only about the
present, but also about the past and future \cite{Tansel3},
\cite{Tsotras1996}. In questions directed to temporal databases (e.g.\
\pref{intro:1} -- \pref{intro:3}), it becomes crucial for \nldb{s} to
interpret correctly temporal linguistic expressions.
\begin{examples}
\item What was the salary of each engineer while ScotCorp was building
bridge 5? \label{intro:1}
\item Did anybody leave site 4 before the chief engineer had
inspected the control room? \label{intro:2}
\item Which systems did the chief engineer inspect on Monday after the
auxiliary generator was in operation? \label{intro:3}
\end{examples}

Temporal database systems currently exist only in the form of
research prototypes (e.g.\ \cite{Boehlen1995c}), but we expect that 
commercially available temporal database systems will appear in the near 
future, because: (i) data storage is constantly becoming cheaper, and 
hence retaining
huge quantities of information about the past is becoming more cost effective,
(ii) there is a strong demand for tools to store and manipulate
time-dependent data (e.g.\ in stock market and data mining applications),
and (iii) research on temporal databases is maturing, to the extent
that consensus proposals are beginning to appear \cite{tdbsglossary},
\cite{TSQL2book}. It is, therefore, desirable to investigate how
interfaces to temporal database systems can be built.

Typically, \nldb{s} have used an
intermediate representation language (usually some form of
logic) to encode the meanings of natural language requests, with the
resulting intermediate language expressions being available for
translation into a suitable database language, often \sqll
\cite{Melton1993}. (The benefits of this include generality,
modularity, portability \cite{Androutsopoulos1995}.)
Building on this architecture, we propose a
framework for constructing \nldb{s} for temporal databases (\nltdb{s})
that draws on ideas from tense and aspect theories (see 
\cite{Comrie,Comrie2} for an introduction), temporal logics
\cite{VanBenthem},\cite{Gabbay1994b},  and temporal databases. We
emphasise that we do not set out to formulate an improved general
tense and aspect theory or temporal logic. Our aim is a more practical
one: to explore how ideas from these areas can be integrated in a
framework that leads to practical \nltdb{s}. We also
concentrate on keyboard-entered English questions that refer to the past or
the present, ignoring the following more difficult issues:
questions about the future (that may require more complex semantic
models), speech input (although our work is in principle 
compatible with the text having been spoken), and requests to update
the database (see \cite{Davidson1983} for an outline of the
difficulties with updates).

\begin{figure}
\hrule
\medskip
\begin{center}
\includegraphics[scale=.6]{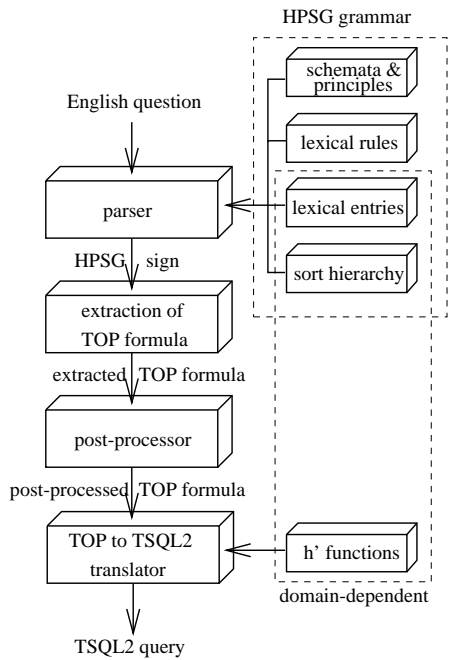}
\caption{Architecture of the prototype \nltdb}
\label{simple-arch-fig}
\end{center}
\hrule
\end{figure}

Ignoring some details, our framework consists of a formal temporal
intermediate representation language, called \topl, an \hpsg grammar
\cite{Pollard2} that maps a wide range of English questions involving
temporal mechanisms to appropriate \topl expressions, and a provably
correct mapping to translate from \topl to \tsql, \tsql being a
recently proposed extension of \sqll \cite{TSQL2book}. To demonstrate
that this framework is workable, we have employed it to implement a
small prototype \nltdb, using \ale \cite{Carpenter1992,Carpenter1994}
and Prolog.\footnote{The prototype \nltdb is available from
\texttt{http://www.dai.ed.ac.uk/groups/nlp}.}  Figure
\ref{simple-arch-fig} shows the architecture of that system.

Each English question is first parsed using the \hpsg grammar,
generating an \hpsg \newt{sign} (a feature-structure that represents
syntactic and semantic properties of the question; signs and the
components of the \hpsg grammar will be discussed in section
\ref{English_to_TOP}).  Multiple signs are generated for questions
that the parser understands to be ambiguous. A \topl formula is then
extracted from each sign and the extracted formula undergoes an
additional post-processing phase (to be discussed in section
\ref{post_process}).

The \topl formulae that are generated at the end of the
post-processing capture what the \nltdb understands to be the possible
readings of the English question. (The prototype \nltdb makes no
attempt to determine which reading the user has in mind.)  These
formulae are then translated into \tsql. (The $h'$
functions of figure \ref{simple-arch-fig} map primitive \topl
expressions to \tsql expressions. They will be discussed in section
\ref{primitive-maps}.)

The prototype \nltdb prints all the
resulting \tsql queries and the corresponding \topl formulae.

The generated \tsql queries would be executed by the underlying
database management system (\dbms) to retrieve the information
requested by the user. A prototype database system, called \timedb,
that supports a version of \tsql now exists.\footnote{\timedb was
developed by Andreas Steiner at ETH Z\"urich. It is available from
\texttt{http://www.cs.auc.dk/}
\texttt{general/DBS/tdb/TimeCenter}. \timedb compiles \tsql statements
into sequences of conventional \sqll statements, which are then
executed by a commercial database system (\textsc{oracle}).}  During
the work that we report here, however, \timedb was not yet
available. As a result of this, our prototype \nltdb has not been
linked to \timedb. Hence, the \tsql queries are currently not
executed, and no answers are produced. We hope that future work will
link the two systems. This task is complicated by the fact that the
two systems adopt different versions of \tsql, since we had to
introduce some modifications to \tsql (see section~\ref{TSQL2}), and
\timedb also introduces some modifications.

The lexicon entries and sort hierarchy of the grammar, and the
definitions of the $h'$ functions of the \topl to \tsql translator are
the only parts of the prototype \nltdb that need to be revised when
configuring the \nltdb for a new database. 

The parser, the grammar, and the formula extractor were
implemented using \ale \cite{Carpenter1992,Carpenter1994}. \ale
provides a chart-parser (which is the one used in the prototype
\nltdb), and a formalism for expressing typed feature-structures,
Prolog-like rules, and unification-based grammars.\footnote{The \ale
grammar of the prototype \nltdb is based on previous \ale encodings of
\hpsg grammars by Penn and Carpenter at Carnegie Mellon University,
and Manandhar and Grover at the University of Edinburgh. We are
grateful to all four of them. \ale is available from
\texttt{http://macduff.andrew.cmu.edu/ale}.}  The post-processor and
the \topl to \tsql translator were implemented in Prolog.

We emphasise that the prototype \nltdb is intended only to demonstrate
that our theoretical framework is implementable. Several modules would
have to be added to the prototype \nltdb, if it were to be used in real-life
applications (see section \ref{further_work}). 

The remainder of this paper is organised as follows: section
\ref{aspectual_classes} discusses some ideas from tense and aspect
theories that are used in our framework; section \ref{TOP_language}
describes \topl, our intermediate representation language; section
\ref{English_to_TOP} presents the \hpsg-based mapping from English to
\topl; section \ref{TSQL2} offers a brief introduction to \tsql; section
\ref{TOP_to_TSQL2} highlights the \topl to \tsql mapping; section
\ref{previous_work} compares our approach to previous proposals on
\nltdb{s}; section \ref{further_work} concludes and proposes
directions for further research.

Although several issues (summarised in section \ref{further_work})
remain to be addressed, we argue that our framework constitutes a
significant improvement over previous proposals on \nltdb{s}.


\section{Aspectual classes} \label{aspectual_classes} 

It is common in tense and aspect theories to classify natural language
expressions or situations described by natural language expressions
into \emph{aspectual classes} (or ``aktionsarten''), a practice often
attributed to \cite{Vendler}. Numerous aspectual taxonomies
have been proposed (see, for example, \cite{Dowty1986}, \cite{Moens2},
\cite{Mourelatos1978}, \cite{Parsons1990}, \cite{Vlach1993}). These
differ in the number and names of the classes, and the nature of the
classified objects. What is common to all of them is that they try
to capture intuitions about the way that native speakers view 
situations, with respect to issues such as whether the situation is
instantaneous or more drawn out, whether the situation involves
an event which is completed
or not, etc.  Although most of this work has been motivated by
the goal of capturing linguistic generalisations, these nuances of
meaning are also relevant to ensuring that a \nltdb operates in
a correct and comprehensible manner.

\subsection{Our aspectual taxonomy}\label{our-aspectual-taxonomy}

For our purposes, we have found a taxonomy that distinguishes between
\emph{states}, \emph{points}, \emph{culminating activities}, and
\emph{activities} to be adequate. Our four classes correspond to ways
of viewing situations that people seem to use when communicating in
English: a situation can be viewed as involving no action (state
view), as an instantaneous action (point view), as an action with a
\emph{climax} (an inherent completion point; culminating activity
view), or as an action with no climax (activity view).  (The use of
``culminating activities'' is derived from \cite{Moens2}.) For
example, \qit{Tank 5 contains oil.} is typically uttered with a state
view in mind, while \qit{Tank 5 exploded.} would be used in most
contexts with a point view. \qit{ScotCorp built a new bridge.} would
be typically uttered with a culminating activity view. In this case,
the building situation is seen as having an inherent completion point
(the point where the entire bridge has been built). If the building
stops before reaching this point, the building situation is considered
incomplete. When a verb is used in the simple past in a sentence
uttered with a culminating activity view, the speaker usually implies
that the climax was reached (the building of the bridge was
completed). In contrast, \qit{Robot 2 moved.} would probably be
uttered with an activity view. In this case, there is no climax (the
movement can stop at any point without being any more or less
complete), and the simple past carries no implication that any climax
was reached. If, however, a particular destination were provided, as
in \qit{Robot 2 moved to the exit.}, the sentence would be uttered
with a culminating activity view, and the climax would be the point
where the robot reaches the exit.

Determining which view the speaker has in mind is important to
understand what the speaker means. For example, when an \qit{at~\dots}
temporal adverbial is attached to a clause uttered with a state view
(e.g.\ \qit{Tank 5 contained oil at 5:00pm.}), the speaker usually
means that the situation of the clause simply holds at the adverbial's
time. Excluding special contexts that do not arise in \nldb{s}, there
is normally no implication that the situation starts or stops holding
at the adverbial's time. In contrast, with clauses uttered with
culminating activity views, \qit{at~\dots} adverbials (if felicitous)
usually specify the time where the situation starts or reaches its
climax (e.g.\ \pref{aspclass:1}). 
\begin{examples}
\item Robot 2 moved to the exit at 5:00pm. \label{aspclass:1}
\end{examples}
Some linguistic markers seem to signal which view the speaker has in
mind. For example, the progressive usually signals a state view (e.g.\
unlike \pref{aspclass:1}, \pref{aspclass:2} is typically uttered with
a state view, and the movement is simply ongoing at 5:00pm).
\begin{examples}
\item Robot 2 was moving to the exit at 5:00pm. \label{aspclass:2}
\end{examples}
Often, however, there are no such explicit markers. The processes
employed in those cases by hearers to determine the speaker's view are
not yet fully understood. In a \nltdb, however, where questions refer
to a restricted domain, reasonable guesses can be made by observing
that in each domain, each verb tends to be associated mainly with one
particular view. Certain agreements about how situations are to be
viewed (e.g.\ that some situations are to be treated as instantaneous
-- point view) will have also been made during the design of the
database. These agreements provide additional clues about the main
view of each verb in the domain. 

More precisely, we adopt the following approach. Whenever the \nltdb
is configured for a new application domain, the base form of each verb
is assigned to one of the four aspectual classes, using criteria (to
be discussed below) intended to determine the main view of each verb in
the particular domain. Aspectual class is treated as a property not
only of verbs, but also of verb phrases, clauses, and sentences. Normally,
all verb forms inherit the aspectual classes of the corresponding base
forms. Verb phrases, clauses, and sentences normally inherit the
aspectual classes of their main verb forms. Some linguistic mechanisms
(e.g.\ the progressive), however, may cause the aspectual class of a
verb to differ from that of the base form, or the aspectual class of a
verb phrase, clause, or sentence to differ from that of its main verb
form. 

In the case of a verb like \qit{to run}, that typically involves a
culminating activity view when used with an expression determining a
specific destination or distance (e.g.\ \qit{to run to the
station/five miles}), but an activity view otherwise, we 
assume that there are two different homonymous verbs \qit{to run}:
one has a culminating activity base form, and requires a complement
determining a specific destination or distance; the other has an
activity base form and requires no such complement. A similar
distinction is introduced with verbs whose aspectual class
depends on whether the verb's object denotes a countable or mass
entity (e.g.\ \qit{to drink a bottle of wine} vs.\ \qit{to drink
wine}) \cite{Mourelatos1978}. Similarly, when a verb can be used in an
application domain with both habitual and non-habitual meanings (e.g.\
\qit{BA737 departs from Gatwick.} (habitually, perhaps every day) vs.\
\qit{BA737 {\rm (actually)} departed from Gatwick five minutes ago.}), we
distinguish between homonyms with habitual and non-habitual
meanings. Our aspectual criteria classify the base forms of habitual
homonyms as states. The aspectual classes of non-habitual homonyms
depend on the verb and the domain.
\label{habitual-ambiguity}

In the rest of this paper, we call verbs whose \emph{base} forms are
classified as states, activities, culminating activities, and points,
\emph{state, activity, culminating
activity, and point verbs} respectively.  This does not rule out
other (non-base) forms of these verbs belonging to other classes.

\subsection{Aspectual criteria}\label{aspect-criteria}

It is important to realise that our aspectual classes play a purely formal
role, in the sense that their whole purpose is to allow the systematic
translation of English into our semantic representation, in a way that
achieves plausible effects within a \nltdb. Despite our attempts, in
section \ref{our-aspectual-taxonomy} above, to provide an
informal meaning for our classes, they are not accessible to
linguistic intuition.  Verbs are located in the classification scheme
using particular criteria, which we outline below, not by considering
whether or not they appear to meet some semantic condition suggested
by the class name. For example, we have found it convenient to group
``habitual'' usage with ``states'', but this does not embody an empirical
claim that ``habitual'' sentences do not describe actions.

The criteria we use to classify the base verb forms in each
application domain are three, and they are applied
sequentially. First, the \emph{simple present criterion} is used: if
the simple present of a verb can be used in the particular domain in
single-clause questions with non-futurate meanings, it is
a state verb; otherwise, it is a point, activity, or
culminating activity verb. For example, in domains where \pref{aspclass:3}
is possible, \qit{to contain} is a state verb.
\begin{examples}
\item Does any tank contain oil? \label{aspclass:3}
\end{examples}
The simple present is sometimes used to refer to scheduled situations.
For example, \pref{aspclass:5} could refer to a particular scheduled
assembling. We consider this meaning of \pref{aspclass:5} a futurate
one. Hence, \pref{aspclass:5} does not constitute evidence that
\qit{to assemble} is a state verb.
\begin{examples}
\item When does J.Adams assemble engine 5? \label{aspclass:5}
\end{examples}
As explained in section~\ref{habitual-ambiguity},
when verbs have both habitual and
non-habitual uses, we distinguish between habitual and non-habitual
homonyms. Excluding scheduled meanings, \pref{aspclass:6} can only
have a habitual meaning, i.e.\ it can only involve the habitual
homonym of \qit{to land}. Therefore, in domains where
\pref{aspclass:6} is possible, the habitual \qit{to land} is 
a state verb. \pref{aspclass:6} does not constitute evidence that the
non-habitual \qit{to land} is a state verb.
\begin{examples}
\item Which flight lands on runway 2? \label{aspclass:6}
\end{examples}

Next, the \emph{point criterion} is applied to distinguish
point verbs from activity or culminating activity verbs. The
point criterion is based on the fact that some verbs describe kinds of
situations that are always modelled in the database as
instantaneous. If a verb describes situations of this kind, it is
a point verb; otherwise it is an activity or culminating
activity verb. Many of our illustrative examples come from a hypothetical 
airport domain (described in \cite{Androutsopoulos1996}). 
In that domain, we assume the database does not distinguish between the
times at which a flight starts or stops entering an airspace sector:
entering a sector is modelled as instantaneous. Also, in our airport
domain \qit{to enter} is used only to refer to flights entering
sectors. Consequently, in that domain \qit{to enter} is a
point verb. (If \qit{to enter} were also used to refer to groups of
passengers entering planes, and if situations of this kind were
modelled as non-instantaneous, one would have to distinguish two
homonymous \qit{to enter}: one used with flights entering sectors, and
one with groups of passengers entering planes. Only the the first
homonym would be a point verb.)

Finally, the \emph{imperfective paradox criterion} is applied to
decide if the remaining verbs are activity or culminating
activity ones. The criterion is based on the ``imperfective paradox''
\cite{Dowty1977}, \cite{Lascarides}. Yes/no questions containing the
past continuous and the simple past of the verbs (e.g.\
\pref{aspclass:7} -- \pref{aspclass:10}) are considered. (The past
continuous questions must not be read with futurate meanings, e.g.\
\pref{aspclass:7} must not be taken to ask if John was \emph{going}
to run.) If an affirmative answer to the past continuous question
implies an affirmative answer to the simple past question (as in
\pref{aspclass:7} -- \pref{aspclass:8}), the verb is an
activity verb; otherwise (as in \pref{aspclass:9} -- \pref{aspclass:10}),
it is a culminating activity verb. (\pref{aspclass:9} does not
imply \pref{aspclass:10}: John may have abandoned the building before
completing the house.)
\begin{examples}
\item Was John running? \label{aspclass:7}
\item Did John run? \label{aspclass:8}
\item Was John building a house? \label{aspclass:9}
\item Did John build a house? \label{aspclass:10}
\end{examples}
We show below how some base verb forms are classified in our
airport domain.
\begin{description}
\item[states:] contain, be (non-auxiliary), arrive (habitually),
depart (habitually), service (habitually)
\item[activities:] circle (waiting permission to land), taxi (no
destination), queue (for runway)
\item[culminating activities:] land, take off, inspect, board, service
(actually), taxi (to destination)
\item[points:] enter, start, begin, stop, finish,
leave, arrive (actually), depart (actually)
\end{description}
We stress that the classification depends on the uses of the
verbs in the particular domain, and on how the situations of the verbs
are modelled in the database. For example, our hypothetical
airport database models
arrivals and departures as instantaneous, and hence the corresponding
verbs are classified as point verbs. If arrivals and departures
were not modelled as instantaneous, the corresponding verbs would
probably be classified as culminating activity ones.


\section{The TOP language} \label{TOP_language} 

We now turn to \topl, our intermediate representation language.
Although we avoid referring to it as a ``logic'' (on the grounds that
it has no proof theory), it is in many respects similar to, and
closely based on, traditional logics, and it has a model theory
much as one would define for a genuine temporal logic. However, it
is far from being a full logic, as the emphasis in its design has
been on reflecting natural language semantics rather than 
covering inferential phenomena. A formal summary of \topl is given
in the appendix; although this is rather terse owing to space limitations,
a fuller account can be found in chapter 3 of \cite{Androutsopoulos1996}.

\topl contains constants and variables, and atomic formulae are
constructed by applying predicate symbols to arguments. Formulae can
be conjoined, but there are currently no negation or disjunction
connectives. (We did not need these for the linguistic phenomena that
we focused on. We plan to include these in future \topl versions.)
Similarly, there are currently no quantifier symbols, with variables
essentially being treated as existentially quantified.  The most
relevant constructs for the current discussion are the temporal
operators (which were influenced by those of \cite{Crouch2}). For
example, \pref{top:1} is represented as \pref{top:2}. Roughly
speaking, the \past operator (introduced by the verb tense) requires
$contain(tk2, water)$ to be true at some past time $e^v$ (the ``$^v$''
suffix marks variables), and the \at operator (introduced by the
adverbial) requires that time to fall within \textit{1/10/95}. (Dates
are shown in the day/month/year format.)
\begin{examples}
\item Did tank 2 contain water (some time) on 1/10/95? \label{top:1}
\item $\at[\mbox{\textit{1/10/95}}, \past[e^v, contain(tk2, water)]]$
\label{top:2} 
\end{examples}
As we are not primarily interested in the structure of referring expressions,
we will make the simplifying assumption that particular items in the domain
such as \qit{tank 2} are associated with simple constants such as
$tk2$ rather than some more complex
logical expression. The prototype \nltdb currently requires \qit{tank 2} to
be typed as a single word, which is associated in the lexicon with $tk2$.
Notice that $tk2$ is a \topl constant, {\em not} a database value.

\subsection{TOP basics} \label{TOP_basics}

\topl assumes that time is linear, discrete, and bounded
\cite{VanBenthem}.  Following the Reichenbachian tradition
\cite{Reichenbach}, \topl formulae are evaluated with respect to more
than one times: $st$ (\emph{speech time}, the time where the question
is submitted to the \nltdb), $et$ (\emph{event time}, the time where
the situation denoted by the formula holds), and $lt$
(\emph{localisation time}, a temporal window within which the event
time must be placed (cf.~\cite{Crouch2}; this is different from
Reichenbach's ``reference time'' and similar to the ``location time''
of \cite{Kamp1993}). While $st$ is always a time-point, $et$
and $lt$ are generally periods, and hence \topl
can be considered period-based. (To avoid conflicts with \tsql
terminology, we use the term ``period'' to refer to what temporal
logicians usually call ``intervals'', i.e.\ convex sets of
time-points.)

Although the aspectual classes of natural language expressions affect
how these expressions are represented in \topl, it is not always
possible to tell the aspectual class of a natural language expression
by examining its \topl representation. The approach here is
different from those of \cite{Dowty1977}, \cite{Dowty1986},
\cite{Kent}, and \cite{Lascarides}, where aspectual class is a
property of formulae or denotations of formulae. 

Returning to \pref{top:1}, the answer will be affirmative if
\pref{top:2} evaluates to true. When evaluating \topl formulae, $lt$
initially covers the entire time axis, though it may be narrowed down
subsequently by temporal operators.  In \pref{top:2}, the \at operator
narrows $lt$ to its intersection with 1/10/95. This causes $lt$ to
become 1/10/95. The \past operator then narrows $lt$ to its
intersection with $[t_{first}, st)$, where $t_{first}$ is the earliest
time-point. (In this particular example, if 1/10/95 is entirely in the
past, the \past operator does not restrict $lt$ any further.) The
overall formula evaluates to true iff there is an event time period
$et$, where $contain(tk2, water)$ is true, such that $et \subper
lt$. ($p_1$ is a subperiod of $p_2$, written $p_1 \subper p_2$, iff
$p_1, p_2$ are periods and $p_1 \subseteq p_2$.)

Although here we provide only informal descriptions of the semantics
of \topl formulae, \topl has formally defined model-theoretic
semantics (see Appendix \ref{top-definitions} and 
{\cite{Androutsopoulos1996}). Among other things, a
\topl model (which is ultimately defined in terms of database
constructs), specifies the maximal event time periods where the
situations denoted by \topl predicates hold (e.g.\ in the case of
$contain(tk2, water)$, it specifies the maximal periods where tank 2
contained water). \topl predicates are always \emph{homogeneous}. This
means that if a predicate is true at an event time $et$, it is also
true at any event time $et' \subper et$. (Various versions of
homogeneity have been used in \cite{Allen1984}, \cite{Kent},
\cite{Lascarides}, \cite{Richards}, and elsewhere.) In our example, if
tank 2 contained water from 29/9/95 to 2/10/95, $contain(tk2, water)$
is true at any event time that is a subperiod of that
period. Hence, there will be an $et$ where $contain(tk2, water)$ is
true, such that $et$ is also a subperiod of 1/10/95 (the $lt$), and
the answer to \pref{top:1} will be affirmative.

\subsection{Interrogative quantifiers} \label{top:interrogs}

In \pref{top:2}, the semantics of the \past operator bind the
variable $e^v$ to $et$ (the event time, where tank 2 contained water). The
$e^v$ argument of the \past operator is useful, for example, in
time-asking questions like \pref{top:3}, expressed as
\pref{top:4}. The interrogative quantifier ($?_{mxl}$) of \pref{top:4}
reports the \emph{maximal} past $et$s where $contain(tk2, water)$ is
true, such that $et$ is a subperiod of 1/10/95.
\begin{examples}
\item When on 1/10/95 was tank 2 empty? \label{top:3}
\item $?_{mxl}e^v \; \at[\mbox{\textit{1/10/95}}, \past[e^v, empty(tk2)]]$ 
   \label{top:4}
\end{examples}

A simpler version of the interrogative quantifier is used in non-time
asking questions like \pref{top:5}, expressed as \pref{top:6}.
\begin{examples}
\item What did tank 2 contain on 1/10/95 at 5:00pm? \label{top:5}
\item $?w^v \begin{array}[t]{l}
            \partop[\mathit{5\_00pm}^g, f^v] \land
            \at[\mbox{\textit{1/10/95}},  
            \at[f^v, \past[e^v, contain(tk2, w^v)]]]
            \end{array}$ \label{top:6}  
\end{examples}
The $\mathit{5\_00pm}^g$ in \pref{top:6} denotes the set of all
5:00pm-periods (5:00:00-59 on 2/2/95, 5:00:00-59 on 3/2/95,
etc.). The ``$^g$'' suffix stands for ``gappy partitioning'': the
set of all 5:00pm-periods is gappy, in the sense that the union of its
elements does not cover the entire time-axis.
$\partop[\mathit{5\_00pm}^g, f^v]$ means that $f^v$ must be a
5:00pm-period. The two \at operators cause $lt$ to become the
particular 5:00pm-period of 1/10/95. $e^v$, the event time where
$contain(tk2, w^v)$ holds, must fall in the past and within $f^v$.

\subsection{More on At, Before, and After}

Two additional operators, \before and \after, are used to express
adverbials like \qit{before 1/10/95} or \qit{after 5:00pm}. Like \at,
these operators narrow $lt$. The first arguments of \at, \before, and
\after can also be formulae. This form of the operators is used when
expressing subordinate clauses introduced by \qit{while},
\qit{before}, or \qit{after} respectively. In \pref{top:10}, which
represents one possible reading of \pref{top:9}, the \at operator causes
$\past[e2^v, free(run2)]$ to be evaluated with respect to a new
$lt'$. $lt'$ is the intersection of the original $lt$ (the whole
time-axis) with a maximal event time period $e1^v$ where
$\past[e1^v, circling(ba737)]$ is true. \pref{top:10} evaluates to
true iff there is a past event time $e2^v \subper lt'$ where
$free(run2)$ is true.
\begin{examples}
\item Was runway 2 free while BA737 was circling? \label{top:9}
\item $\at[\!\!\!\begin{array}[t]{l}\past[e1^v, circling(ba737)],
 \past[e2^v, free(run2)]]
           \end{array}$ \label{top:10} 
\end{examples}

\subsection{The Fills operator}

\pref{top:10} represents the reading of \pref{top:9} whereby runway 2
must have been free \emph{some} time while BA737 was circling. There
is, however, a stricter reading, that requires runway 2 to have been
free \emph{throughout} BA737's circling.
(Some speakers report another, preferred, reading in which 
the runway had to be free for \emph{most} of the time that 
BA737 was circling. As a simplification, we will not handle that
dialect variant, which would be semantically complicated.)
Readings of this kind are
very common when states combine with adverbials or clauses that are
understood as specifying non-instantaneous periods. For example,
\pref{top:12} has a reading where tank 2 must have been empty
\emph{some} time on 1/10/95, and a stricter (and probably preferred)
one where the tank must have been empty \emph{throughout} that day.
\begin{examples}
\item Was tank 2 empty on 1/10/95? \label{top:12}
\end{examples}
For simplicity, the stricter readings are currently ignored in our
English to \topl mapping (and the rest of this paper). These readings,
however, \emph{can} be expressed in \topl using the \fills operator,
which requires the event time to cover exactly the localisation
time. For example, the stricter reading of \pref{top:9} can be
expressed as \pref{top:11}. The \fills operator of \pref{top:11}
requires $e2^v$ (event time where runway 2 was free) to be set to
$lt'$ (maximal past period where BA737 was circling). The stricter
reading of \pref{top:12} is expressed as \pref{top:13}.
\begin{examples}
\item $\at[\!\!\!\begin{array}[t]{l}\past[e1^v, circling(ba737)],
                              \past[e2^v, \fills[free(run2)]]]
           \end{array}$ \label{top:11}
\item $\at[\mbox{\textit{1/10/95}}, \past[e^v, \fills[empty(tk2)]]]$ 
\label{top:13}
\end{examples}

\subsection{The Culm operator} \label{culm_operator}

In the case of activity, state, or point verbs (e.g.\ \qit{to circle},
\qit{to contain}, and \qit{to depart} in our airport domain),
progressive and non-progressive forms receive the same \topl
representations. For example, both \pref{top:14} and \pref{top:15} are
represented as \pref{top:16}, which requires an $et$ where BA737 was
circling to exist in the past. (Progressive forms of state verbs are
typically unfelicitous, but for our purposes it is harmless to allow
them. The use of point verbs in the progressive, as in
\pref{top:16.1}, usually signals that the user is not aware that the
situation of the verb is modelled as instantaneous. A warning to the
user should be generated in the latter case. Our prototype \nltdb,
however, currently does not generate such messages.)
\begin{examples}
\item Was BA737 circling? \label{top:14}
\item Did BA737 circle? \label{top:15}
\item $\past[e^v, circling(ba737)]$ \label{top:16}
\item Was BA737 departing at 5:00pm? \label{top:16.1}
\end{examples}
With culminating activity verbs (e.g.\ \qit{to inspect} in
the airport domain), the representation of the non-progressive
contains an additional \culm operator. \pref{top:17} and \pref{top:19}
are represented as \pref{top:18} and \pref{top:20} respectively. The
\culm operator requires $et$ to begin at the start-point of the
earliest maximal period where the situation of its argument
holds (the argument of \culm is always an atomic formula), and to end at the
end-point of the latest of those maximal periods. Furthermore, the
situation of the predicate must reach its climax at the end-point of
$et$.
\begin{examples}
\item Was J.Adams inspecting BA737? \label{top:17}
\item $\past[e^v, inspecting(ja, ba737)]$ \label{top:18}
\item Did J.Adams inspect BA737? \label{top:19}
\item $\past[e^v, \culm[inspecting(ja, ba737)]]$ \label{top:20}
\end{examples}

Let us assume, for example, that J.Adams started to inspect BA737 at
9:00pm on 1/10/95, stopped the inspection at 1:00pm, resumed it at
2:00pm, and finished it at 4:00pm on the same day.  The maximal
periods where the situation of $inspecting(ja, ba737)$ holds are from
9:00pm to 1:00pm and from 2:00pm to 4:00pm on 1/10/95. Let us call
$p_{mxl1}$ and $p_{mxl2}$ these maximal periods. \pref{top:18}
requires an $et$ where the predicate holds (i.e.\ a subperiod of
$p_{mxl1}$ or $p_{mxl2}$) to be a subperiod of $[t_{first}, st)$ (the
$lt$). That is, there must be some past time where J.Adams was
inspecting BA737. \pref{top:20}, on the other hand, would, with
the database values assumed here, require $et$ to
be the period from 9:00pm to 4:00pm on 1/10/95, the inspection to
reach its climax at the end of $et$, and $et \subper [t_{first},
st)$. That is, the entire inspection must be in the past. If the
questions were submitted at 3:00pm on 1/10/95, \pref{top:18} would be
true, but \pref{top:20} would be false. This accounts for the fact
that \pref{top:19} requires the inspection to have been completed,
while \pref{top:17} does not. 

We note that our approach does not run into problems with the
``imperfective paradox'' (\pref{top:20} implies \pref{top:18}, but not
the opposite; cf.\ the logic of \cite{Crouch2}), and this is achieved
without invoking branching-time or possible worlds (cf.\
\cite{Dowty1977}, \cite{Kent}), that are difficult to model in
practical temporal database systems.

\subsection{Episode identifiers}

With culminating activity verbs, we often use an extra
\emph{episode-identifying} argument (the variable $ep^v$ in
\pref{top:22}; $ep^v$ should also, for correctness, be present in
\pref{top:18} and \pref{top:20}). This is needed in sentences like
\pref{top:21}, if J.Adams was involved in several different
inspections of BA737. (We treat assertions like \pref{top:21} as
yes/no questions; variables can be thought of, informally, as existentially
quantified.)
\begin{examples}
\item J.Adams inspected BA737 on 1/10/95. \label{top:21}
\item $\!\!\!\begin{array}[t]{l}
        \at[\mbox{\textit{1/10/95}}, \past[e^v,
        \culm[inspecting(ep^v, ja, ba737)]]]
       \end{array}$ \label{top:22} 
\end{examples}
Let us assume, for example, that apart from the inspection of 1/10/95,
J.Adams was also involved in another inspection of BA737, that started
at 11:00pm on 1/1/95, and stopped (without ever reaching its
completion) at 4:00am on 2/1/95. Let us call $p_{mxl3}$ this period.
We use the episode-identifying argument to distinguish between the two
inspections: $inspecting(ep1, ja, ba737)$ is true for $et \subper
p_{mxl1}$ or $et \subper p_{mxl2}$, while $inspecting(ep2, ja, ba737)$
is true for $et \subper p_{mxl3}$.

The preferred reading of \pref{top:21} is that an entire inspection
(from start to completion) of BA737 by J.Adams occurred within
1/10/95. Without the episode-identifying argument, this reading is not
captured correctly by \pref{top:22}: \pref{top:22} would require $et$
to extend from the beginning of $p_{mxl3}$ (on 1/1/95) to the end of
$p_{mxl2}$ (on 1/10/95), and $et$ to be a subperiod of 1/10/95. Since this
is impossible, the answer would (incorrectly) be false. With $ep^v$,
in contrast, \pref{top:22} requires $et$ to extend from the beginning
of the earliest maximal period that corresponds to a particular value
of $ep^v$ ($ep1$ or $ep2$) to the end of the latest maximal period
that corresponds to the \emph{same} $ep^v$ value. $et$ must also be a
subperiod of 1/10/95, and the inspection that corresponds to
$ep^v$'s value must reach its climax at the end of $et$. For $ep^v =
ep2$, all these conditions are satisfied, and hence the answer would
(correctly) be affirmative.

\subsection{The For operator}

Duration adverbials are represented using the \for operator. For
example, \pref{top:29} is mapped to \pref{top:30}. $day^c$ and
$minute^c$ represent the sets of all day- and minute-periods
respectively. (The ``$^c$'' suffix stands for ``complete partitioning'': the
sets of all day- and minute-periods are complete, in the sense that the
union of each set's elements is the entire time-axis.) The \for
operator of \pref{top:30} requires $et$ (the time of the circling) to
cover 40 consecutive minutes. 
\begin{examples}
\item On which day was BA737 circling for forty minutes? \label{top:29}
\item $?d^v \!\begin{array}[t]{l}
       \partop[day^c, d^v] \land \at[d^v, 
       \for[min^c, 40, \past[e^v, circling(ba737)]]
       \end{array}$ \label{top:30}
\end{examples}

\subsection{The Begin and End operators}

Verbs like \qit{to start}, \qit{to stop}, and \qit{to finish} are
expressed using the \lbegin and \lend operators.  \pref{top:27} and
\pref{top:25} are mapped to \pref{top:28} and \pref{top:26}
respectively. \pref{top:28} requires the end-point of a maximal period
where the inspection is ongoing to be located at some past
5:00pm-minute. \pref{top:26} requires the \emph{completion} of an
inspection to fall within a past 5:00pm-minute.
\begin{examples}
\item J.Adams stopped inspecting BA737 at 5:00pm. \label{top:27}
\item $\!\!\!\begin{array}[t]{l}
       \partop[$\textit{5:00pm}$^g, f^v] \land \at[f^v, \past[e^v,
       \lend[inspecting(ep^v, ja, ba737)]]]
       \end{array}$ \label{top:28} 
\item J.Adams finished inspecting BA737 at 5:00pm. \label{top:25}
\item $\!\!\!\begin{array}[t]{l}
       \partop[$\textit{5:00pm}$^g, f^v] \land \at[f^v, \past[e^v,
       \lend[\culm[inspecting(ep^v, ja, ba737)]]]]
       \end{array}$ \label{top:26} 
\end{examples}

\subsection{The Pres and Perf operators}\label{pres-and-perf}

The past perfect is expressed using a \past followed by a \perf
operator. For example, \pref{top:30b} is mapped to \pref{top:31}. The
\perf operator introduces a new event time (denoted by $e2^v$ in
\pref{top:31}), that has to precede the original one
($e1^v$). \pref{top:31} requires the time of the departure ($e2^v$) to
precede another past time ($e1^v$). The latter corresponds to
Reichenbach's \emph{reference time} \cite{Reichenbach}, a time
from which the departure is viewed.  With culminating
activity verbs, a \culm operator is also inserted (e.g.\ \pref{top:32}
receives \pref{top:33}).
\begin{examples}
\item BA737 had departed. \label{top:30b}
\item $\past[e1^v, \perf[e2^v, depart(ba737)]]$ \label{top:31}
\item Had J.Adams inspected BA737? \label{top:32}
\item $\!\!\!\begin{array}[t]{l}
       \past[e1^v, \perf[e2^v, 
       \culm[inspecting(ep^v, ja, ba737)]]]
      \end{array}$ \label{top:33} 
\end{examples}
A sentence like \pref{top:34} has two readings: \qit{on 1/1/95} may
specify the reference time, or it may refer directly to the
inspection. (In the latter case, \pref{top:34} is similar to \qit{Did
J.Adams inspect BA737 on 1/1/95?}, except that it creates an
impression of a longer distance between the inspection and the
present.) We represent the two readings as \pref{top:35} and
\pref{top:36} respectively.
\begin{examples}
\item Had J.Adams inspected BA737 on 1/1/95? \label{top:34}
\item $\!\!\!\begin{array}[t]{l}
       \at[\mbox{\textit{1/1/95}}, \past[e1^v, \perf[e2^v, 
       \culm[\mathit{inspecting}(ep^v, ja, ba737)]]]
       \end{array}$ \label{top:35}
\item $\!\!\!\begin{array}[t]{l}
       \past[e1^v, \perf[e2^v, \at[\mbox{\textit{1/1/95}},
       \culm[\mathit{inspecting}(ep^v, ja, ba737)]]]
       \end{array}$ \label{top:36}
\end{examples}
\pref{top:35} requires the reference time ($e1^v$) to be a subperiod
of 1/1/95, and to follow an entire inspection of BA737 by
J.Adams. Apart from introducing a new $et$, the \perf operator also
resets $lt$ to the entire time-axis. Hence, in \pref{top:35}, $e2^v$
does not need to be a subperiod of 1/1/95. In \pref{top:36}, the \at
operator narrows $lt$ after the \perf operator has
reset it. This requires the inspection to be located within 1/1/95.

It has often been claimed (e.g.\ \cite{Blackburn1994}, \cite{Moens2},
\cite{Vlach1993}) that the English present perfect asserts that some
consequence of a past situation holds at the present. For example,
\pref{top:40} seems to imply that the engine is still on fire, or that
it was damaged by the fire and has not been repaired. (A similar
connection may exist between the past situation and the reference time
in some uses of the past perfect.) This implication does not seem to
be present (at least not as strongly) in \pref{top:41}.
\begin{examples}
\item Engine 5 has caught fire. \label{top:40}
\item Engine 5 caught fire. \label{top:41}
\end{examples}
Although these claims are intuitively appealing, it is difficult to
see how they could be used in a \nltdb. Perhaps in \pref{top:42} the
\nltdb should check not only that the landing was completed, but also
that some consequence of the landing still holds. But what should this
consequence be? Should the \nltdb check that the plane is still at the
airport, and should the answer be negative if the plane has departed
again? Should the \nldb check that the passengers are still at the
airport? Given this uncertainty, we do not require the past situation
to have present consequences. 
\begin{examples}
\item Has BA737 landed? \label{top:42}
\end{examples}

The present perfect could be expressed using a \pres and a \perf
operator. (The \pres operator simply requires $st$ to be located
within $et$.) For example, \pref{top:42} would be represented as
\pref{top:43}. Given, however, that we do not require the past
situation to have present consequences, there are very few differences
between the present perfect and the simple past for our purposes. For
simplicity, we represent the present perfect in the same way as the
simple past, i.e.\ both \pref{top:42} and \pref{top:44} receive
\pref{top:45}.
\begin{examples}
\item $\pres[\perf[e^v, \culm[landing(ep^v, ba737)]]]$ \label{top:43}
\item Did BA737 land? \label{top:44}
\item $\past[e^v, \culm[landing(ep^v, ba737)]]]$ \label{top:45}
\end{examples}

\subsection{The Ntense operator} \label{ntense_op}

We end the presentation of \topl with the \ntense operator (borrowed
from \cite{Crouch2}). This is useful in sentences like \pref{top:46},
where \qit{president} may refer either to the present (the current
president visited Rome, perhaps before becoming president), or to the
time of the visit (the visitor was the president of that time). In
\pref{top:47}, the \ntense operator requires $p^v$ to be the president
at $st$ (indicated by $now^*$). In contrast, in \pref{top:48} the \ntense requires $p^v$ to
be the president at the time of the visit ($e^v$).
\begin{examples}
\item The president visited Rome in 1991. \label{top:46}
\item $\ntense[now^*, president(p^v)] \; \land$ 
      $\at[y1991, \past[e^v, visiting(p^v, rome)]]$ \label{top:47}
\item $\ntense[e^v, president(p^v)] \; \land$ 
      $\at[y1991, \past[e^v, visiting(p^v, rome)]]$ \label{top:48}
\end{examples}
We discuss \pref{top:46} further in section \ref{nominals_time}.


\section{From English to TOP} \label{English_to_TOP}

The English requests are mapped to \topl formulae using an \hpsg-based
grammar \cite{Pollard2}. An \hpsg grammar consists of \emph{schemata},
\emph{principles}, \emph{lexical rules}, \emph{lexicon entries}, and a
\emph{sort hierarchy}. Our grammar is very
close to that of Pollard and Sag, with the main difference being that we use
\topl expressions instead of situation-theoretic constructs. We have
also added an {\feat aspect} feature and principle to accommodate our
aspectual taxonomy, and lexical signs and rules for verb forms,
temporal adverbials, and other temporal mechanisms.

This section attempts to offer a taste of how our \hpsg-based mapping
from English to \topl works (the full details can be found in
\cite{Androutsopoulos1996}). Readers with previous exposure to modern
unification-based grammars should be able to follow most of the
discussion. Some of the details, however, may be unclear to readers
not familiar with \hpsg.

\subsection{Verb forms}\label{aspect-principle}

In \hpsg, each word and syntactic constituent receives a \emph{sign},
a feature-structure of a particular form, that provides information
about the word or syntactic constituent. \pref{hpsg:1} shows the
(lexical) sign
of the base form of \qit{to inspect} in the airport
domain. (To save space, we
ignore some uninteresting features, and we write ``{\feat ss}''
instead of ``{\feat synsem}''.)
\begin{examples}
\item 
\avmoptions{active}
\begin{avm}
[\avmspan{phon \, \<\fval inspect\>} \\
 ss|loc|cat & [head  & \osort{verb}{
                                 [vform & bse \\
                                  aux   & $-$ ]} \\
                         aspect & culmact \\
                         subj   & \< \feat np$\,_{@1}$ \>  \\
                         comps  & \< \feat np$\,_{@2}$ \> ]\\
 \avmspan{ss|cont \, $inspecting\(ep^v, @1, @2\)$} ]
\end{avm}
\label{hpsg:1}
\end{examples}
According to \pref{hpsg:1}, \qit{inspect} is the base form of a
non-auxiliary verb, it is a culminating activity, and it requires a
noun-phrase as its subject, and one as its object. The base form
introduces a \topl expression of the form $\mathit{inspecting}(ep^v,
\avmbox{1}, \avmbox{2})$, where \avmbox{1} and \avmbox{2} correspond
to the entities denoted by the subject and object respectively. (More
precisely, \avmbox{1} and \avmbox{2} stand for \hpsg
\emph{indices}. In our \hpsg version, indices represent \topl
constants and variables. Our indices have a special feature, not shown
here, that distinguishes constant- from variable-representing
indices.) The {\feat cont} values are actually feature-structures that
represent \topl expressions. For simplicity, however, here we show
directly the corresponding \topl expressions.

The signs for non-base forms of non-auxiliary verbs are generated
automatically from the corresponding base form signs by lexical
rules. For example, the simple past \pref{hpsg:2} is generated from
\pref{hpsg:1} by a lexical rule that inserts a \past and a \culm in
the {\feat cont} of the base sign. The lexical rule also modifies
{\feat phon} for simple past morphology, and changes {\feat vform} to
{\srt fin}\/ (finite form, a form that requires no
auxiliary verb). The simple past signs of state, point, and
activity verbs are generated by a similar rule, which does not insert,
however, a \culm operator.
\begin{examples}
\item 
\avmoptions{active}
\begin{avm}
[\avmspan{phon \, \<\fval inspected\>} \\
 ss|loc|cat & [head  & \osort{verb}{
                                 [vform & fin \\
                                  aux   & $-$ ]} \\
                         aspect & culmact \\
                         subj   & \< \feat np$\,_{@1}$ \>  \\
                         comps  & \< \feat np$\,_{@2}$ \> ]\\
 \avmspan{ss|cont \, $\past\[e^v, \culm\[inspecting\(ep^v, @1, @2\)\]\]$} ]
\end{avm}
\label{hpsg:2}
\end{examples}
When \qit{inspected} combines with its subject and object in
\pref{hpsg:2.1}, the {\feat cont} value of \pref{hpsg:2} becomes
\pref{hpsg:2.2}, a value which is inherited by the sign of the overall
sentence. Roughly speaking, the \topl representation of a sentence is
the {\feat cont} value of the sentence's sign. (We will refine this
statement in the following paragraphs.)
\begin{examples}
\item J.Adams inspected BA737. \label{hpsg:2.1}
\item $\past[e^v, \culm[inspecting(ep^v, ja, ba737)]]$ \label{hpsg:2.2}
\end{examples}

Present participle signs are the same as the corresponding base ones,
except that their {\feat vform} is {\srt prp}. In \pref{hpsg:2.10},
for example, apart from its {\feat vform}, the sign for
\qit{inspecting} is the same as \pref{hpsg:1} (i.e.\ \qit{inspecting}
is also a culminating activity form). \pref{hpsg:2.11} shows
the sign of \qit{was} that is used in past continuous forms.
\begin{examples}
\item J.Adams was inspecting BA737. \label{hpsg:2.10}
\item
\avmoptions{active}
\begin{avm}
[\avmspan{phon \, \<\fval was\>} \\
 ss|loc|cat & [head  & \osort{verb}{
                              [vform & fin \\
                               aux   & $+$ ]} \\
                      aspect & state \\
                      subj & \<@1\> \\
                      \avmspan{{\feat comps} \, \<\feat 
                                 vp[subj & \<@1\> \\
                                    vform & {\fval prp}]:@2
                               \>} ]\\ 
 \avmspan{ss|cont \, $\past\[e^v, @2\]$}]
\end{avm}
\label{hpsg:2.11}
\end{examples}
According to \pref{hpsg:2.11}, \qit{was} requires a present participle
verb phrase complement (e.g.\ \qit{inspecting BA737}). The subject of
\qit{was} (\avmbox{1}) is the same as the subject of the
complement. The {\feat cont} of \qit{was} is also the same as the
{\feat cont} of the verb phrase complement (\avmbox{2}), but with an
additional \past operator. In \pref{hpsg:2.10}, when \qit{was}
combines with its complement (\qit{inspecting BA737}) and its subject
(\qit{J.Adams}), the {\feat cont} of \pref{hpsg:2.11} becomes
\pref{hpsg:2.12}. The sign of the overall \pref{hpsg:2.10} inherits
the {\feat cont} of \pref{hpsg:2.11} (i.e.\ \pref{hpsg:2.10} is mapped
to \pref{hpsg:2.12}). 
\begin{examples}
\item $\past[e^v, inspecting(ep^v, ja, ba737)]$ \label{hpsg:2.12}
\end{examples}
The propagation of the {\feat aspect} feature is controlled by our
{\feat aspect} principle: syntactic constituents inherit the {\feat
aspect} of their head-daughter (the \qit{was} in \qit{was
inspecting}), except if the head-daughter combines with an
adjunct-daughter (a modifier), in which case the mother syntactic
constituent inherits the {\feat aspect} of the adjunct-daughter (an
example of the latter case will be given in section
\ref{temp_advs}). In \pref{hpsg:2.10}, the \qit{was inspecting} (and
then the overall sentence) inherits the {\srt state}\/ {\feat aspect}
of \qit{was}. In effect, the progressive has caused an aspectual shift
to state (\qit{inspecting} was a culminating activity). Along with our
treatment of \qit{at~\dots} temporal adverbials (discussed in section
\ref{temp_advs} below), this shift accounts for the fact that in
\qit{J.Adams was inspecting BA737 at 5:00pm.} the inspection was
probably simply ongoing at 5:00pm (cf.\ \qit{J.Adams inspected BA737
at 5:00pm.}). A similar shift is assumed in \cite{Dowty1986},
\cite{Moens2}, \cite{Vlach1993}, and elsewhere.

\subsection{Habituals and non-habituals}

In the case of verbs with multiple homonyms (section
\ref{aspectual_classes}), the lexicon provides multiple signs for the
base forms. \pref{hpsg:4} and \pref{hpsg:3}, for example, correspond
to the habitual and non-habitual homonyms respectively of \qit{to
service}. (In our airport domain, each flight is habitually (normally)
serviced by particular servicing companies, though sometimes a flight
may be serviced by a company other than its normal one.) 
\begin{examples}
\avmoptions{active}
\item 
\begin{avm}
[\avmspan{phon \, \<\fval service\>} \\
 ss|loc|cat & [head  & \osort{verb}{
                              [vform & bse \\
                               aux   & $-$ ]} \\
                      aspect & state \\
                      subj   & \<\feat np$\,_{@1}$\>  \\
                      comps  & \<\feat np$\,_{@2}$\> ]\\ 
 \avmspan{ss|cont \, $hab\_serv\(@1, @2\)$}]
\end{avm}
\label{hpsg:4}
\item 
\begin{avm}
[\avmspan{phon \, \<\fval service\>} \\
 ss|loc|cat & [head  & \osort{verb}{
                              [vform & bse \\
                               aux   & $-$ ]} \\
                      aspect & culmact \\
                      subj   & \<\feat np$\,_{@1}$\>  \\
                      comps  & \<\feat np$\,_{@2}$\> ]\\ 
 \avmspan{ss|cont \, $servicing\(ep^v, @1, @2\)$}]
\end{avm}
\label{hpsg:3}
\end{examples}
\pref{hpsg:4} classifies the habitual homonym as a state verb, and
introduces a predicate of the form $hab\_serv(\avmbox{1},
\avmbox{2})$, which is intended to hold at times where \avmbox{1} (the
servicer) has the habit of servicing \avmbox{2} (the
flight). \pref{hpsg:3} classifies the non-habitual homonym as a
culminating activity verb, and introduces a predicate of the form
$servicing(ep^v, \avmbox{1}, \avmbox{2})$, which is intended to hold
at times where \avmbox{1} is actually servicing \avmbox{2}.

Simple present signs are generated by a lexical rule that requires the
base form to be a state. Thus, no simple present signs are generated
for activity, culminating activity, or point verbs. This
reflects the fact that (ignoring futurate meanings, which we do not
address) these verbs are unlikely to be used in the simple present in a
\nldb context. For example, it is unlikely that \pref{hpsg:5} would 
refer to a currently ongoing service (i.e.\ with a reading
that involves the non-habitual culminating activity verb); 
the present continuous would have been used
instead. \pref{hpsg:5} typically has a habitual meaning (that involves
the habitual state verb).
\begin{examples}
\item Which company services BA737? \label{hpsg:5}
\end{examples}
In \pref{hpsg:5}, this arrangement assigns only a habitual sign to
\qit{services} (the habitual sign is similar to \pref{hpsg:4}, with an
additional \pres operator). This causes \pref{hpsg:5} to receive only
\pref{hpsg:6}, which asks for the current habitual servicer of
BA737.
\begin{examples}
\item $?c^v \; \ntense[now^*, company(c^v)] \; \land$
      $\pres[hab\_serv(c^v, ba737)]$ \label{hpsg:6}
\end{examples}
In contrast, the simple past lexical rules generate both habitual and
non-habitual simple past signs (derived from \pref{hpsg:4} and
\pref{hpsg:3}). Hence, \pref{hpsg:7} receives both \pref{hpsg:8} and
\pref{hpsg:9}. These correspond to two possible readings of
\pref{hpsg:7}: \pref{hpsg:8} asks for the 1990 habitual servicers of
BA737, while \pref{hpsg:9} asks for companies that actually serviced
(possible just once) BA737 in 1990. (The role of the \ntense operators
will be discussed in section \ref{nominals_time}.)
\begin{examples}
\item Which companies serviced BA737 in 1990? \label{hpsg:7}
\item $?c^v \; \ntense[e^v, company(c^v)] \land \at[y1990,$ 
      $\past[e^v, hab\_serv(c^v, ba737)]]$ \label{hpsg:8}
\item $?c^v \; \ntense[e^v, company(c^v)] \land \at[y1990,$ 
      $\past[e^v, \culm[\mathit{servicing}(ep^v, c^v, ba737)]]]$
      \label{hpsg:9}
\end{examples}

\subsection{Temporal adverbials} \label{temp_advs}

Let us now consider the adverbial of \pref{hpsg:102}. The lexicon
provides multiple signs for prepositions that introduce temporal
adverbials. \pref{hpsg:100} shows the \qit{at} sign that is used when
an \qit{at~\dots} adverbial modifies a state or point expression. (The
{\feat head} of \pref{hpsg:100} is shown separately in
\pref{hpsg:101}.)
\begin{examples}
\item J.Adams was inspecting BA737 at 5:00pm. \label{hpsg:102}
\item
\avmoptions{active}
\begin{avm}
[\avmspan{phon \; \<\fval at\>} \\
 ss|loc|cat & [head   & {\rm \pref{hpsg:101}} \\
               subj   & \<\>  \\
               comps  & \<\feat np$\,_{min\_per@1}$\> \\
               aspect & point]\\ 
 \avmspan{ss|cont \, $\at\[@1, @2\]$}]
\end{avm}
\label{hpsg:100}
\item 
\avmoptions{active}
\begin{avm}
\osort{prep}{
[mod|loc|cat|aspect & {\fval state $\lor$ point} \\
 \avmspan{mod \; \feat s[vform {\fval fin}]:@2 $\lor$ 
                 \feat vp[vform {\fval psp}]:@2}]}
\end{avm}
\label{hpsg:101}
\end{examples}
According to \pref{hpsg:100}, \qit{at} requires as its complement a
noun phrase that denotes a minute-period.\footnote{
For readers familiar with
\hpsg, {\srt min\_per}\/ is a subsort of {\srt index}. In our \hpsg
version, indices are divided into subsorts that correspond to types of
world entities. These subsorts are used to enforce selectional
restrictions.}
Requiring the complement of \pref{hpsg:100} to denote a
minute-period ensures that \pref{hpsg:100} will not be used with a
noun phrase complement like \qit{Monday} or \qit{gate 2}. (There are
other \qit{at} signs for uses of \qit{at} in sentences like \qit{BA737
arrived at gate 2}.) The {\feat ss$\mid$loc$\mid$cat$\mid$aspect} of
\pref{hpsg:100} is the aspectual class of the modified expression
\emph{after} attaching the \qit{at~\dots} adverbial. This will be
discussed further below. 

\pref{hpsg:101} (the {\feat head} of \pref{hpsg:100}) shows that
\pref{hpsg:100} can be used when \qit{at} introduces a 
modifier of a state or point expression. It also requires the
expression being modified to be a finite sentence (a finite verb form
that has combined with its complements and subject) or a
past-participle ({\srt psp}\/) verb phrase. We generally require
temporal adverbials (e.g.\ \qit{at 5:00pm}, \qit{on Monday}) to modify
whole finite sentences (e.g.\ \qit{J.Adams inspected BA737}) rather
than finite verb phrases (e.g.\ \qit{inspected BA737}), because in
\hpsg the latter approach would lead to problems with questions like
\pref{hpsg:102b}: in \pref{hpsg:102b}, \qit{was} combines in one step
with both its subject \qit{J.Adams} and its complement \qit{BA737},
following the head-subject-complement schema of \cite{Pollard2};
hence, there would be no verb phrase constituent (verb that has
combined with its complements but not its subject) to be modified by
\qit{at 5:00pm}. 
\begin{examples}
\item Was J.Adams inspecting BA737 at 5:00pm? \label{hpsg:102b}
\end{examples}
Temporal adverbials are also allowed to modify past-participle verb
phrases (see \pref{hpsg:101}). This is needed to be able to generate
both readings of past perfect sentences like \pref{top:34}. We discuss
this in following paragraphs. 

When \qit{at} combines with \qit{5:00pm} in \pref{hpsg:102}, \qit{at
5:00pm} receives \pref{hpsg:103}.
\begin{examples}
\item
\avmoptions{active}
\begin{avm}
[\avmspan{phon \, \<\fval at, 5:00pm\>} \\
 ss|loc|cat & [head   & {\rm \pref{hpsg:101}} \\
               subj   & \<\>  \\
               comps  & \<\> \\
               aspect & point]\\ 
 \avmspan{ss|cont \, $\at\[@1, @2\]$} \\
 \avmspan{qstore \; \{$\partop\[\mbox{\textit{5:00pm}}^g, @1\]$\}}]
\end{avm}
\label{hpsg:103}
\end{examples}
The {\feat qstore} (quantifier storage) of \pref{hpsg:103} shows that
the \topl variable represented by \avmbox{1} is existentially
quantified over all 5:00pm-periods. ({\feat qstore} is intuitively a
set containing quantifiers along with their restrictions. Currently,
all \topl
variables are implicitly existentially quantified. This is why no
explicit existential quantifier is shown in \pref{hpsg:103}.)

The \qit{at 5:00pm} then attaches to \qit{J.Adams was inspecting
BA737}, and \pref{hpsg:102} receives \pref{hpsg:104}. The principles
of \hpsg cause \pref{hpsg:104} to inherit the {\feat head} of
\pref{hpsg:2.11}, and the {\feat cont} and {\feat qstore} of
\pref{hpsg:103}. Our {\feat aspect} principle 
(section \ref{aspect-principle}) also causes
\pref{hpsg:104} to inherit the point {\feat aspect} of
\pref{hpsg:103}. In effect, the point-specifying adverbial has caused
the originally state sentence to become point. (This aspectual shift
is needed to block some impossible readings in sentences with multiple
temporal adverbials; see \cite{Androutsopoulos1996}.)
\begin{examples}
\item
\avmoptions{active}
\begin{avm}
[\avmspan{phon \, \<\fval J.Adams.,was,inspecting,BA737,at,5:00pm\>} \\
 \avmspan{ss|loc|cat \, [head  & \osort{verb}{
                              [vform & fin \\
                               aux   & $+$ ]} \\
                      aspect & point \\
                      subj & \<\> \\
                      comps & \<\>]}\\ 
 ss|cont & $\at\[@1,$ \\
         & $\past\[e^v, inspecting\(ep^v, ja, ba737\)\]\]$\\
 qstore & \{$\partop\[\mbox{\textit{5:00pm}}^g, @1\]$\}]
\end{avm}
\label{hpsg:104}
\end{examples}

To obtain the \topl translation of \pref{hpsg:102}, the expressions in
the {\feat qstore} of \pref{hpsg:104} have to be added in front of the
{\feat cont} formula. (The variable-representing \avmbox{1} also has
to be replaced by an actual \topl variable.) This leads to
\pref{hpsg:105}, which requires the inspection to have been ongoing at
some 5:00pm-minute. (The reader is reminded that assertions are
treated as yes/no questions. \pref{hpsg:102b} receives the same formula.)
\begin{examples}
\item $\partop[\mbox{\textit{5:00pm}}^g, f^v] \; \land$ 
      $\at[f^v, \past[e^v, inspecting(ep^v, ja, ba737)]]$ \label{hpsg:105}
\end{examples}

These examples also demonstrate an important area in which our
framework needs to be extended. In \pref{hpsg:102b} (or
\pref{hpsg:102}) the user would typically have a particular
5:00pm-minute in mind (e.g.\ the 5:00pm-minute of the current or a
previously mentioned day; see section 5.5.1 of \cite{Kamp1993}). A
temporal anaphora resolution mechanism would be needed to determine
that particular minute from the context, but our framework currently
provides no such mechanism.  The answer to \pref{hpsg:102b} would
therefore be affirmative if J.Adams was inspecting BA737 at \emph{any}
past 5:00pm-minute. A similar mechanism is also needed to cope with
the anaphoric nature of tenses \cite{Webber1988}. \pref{hpsg:106}, for
example, is typically a request to report flights that were circling
at a particular contextually salient past time. In contrast, our
framework currently generates an answer that reports all the flights
that were circling at any past time.
\begin{examples}
\item Which flights were circling? \label{hpsg:106}
\end{examples}

Let us now consider the case where an \qit{at~\dots} adverbial
attaches to a culminating activity. The intended meaning is usually
that the situation either starts or reaches its completion at the time
of the adverbial. The first reading is the preferred one in
\pref{hpsg:107}. The second reading is the preferred one in
\pref{hpsg:108}. 
\begin{examples}
\item J.Adams inspected BA737 at 5:00pm. \label{hpsg:107}
\item BA737 landed at 5:00pm. \label{hpsg:108}
\end{examples}
There is an \qit{at} sign for each reading. (There are also additional
signs for cases where \qit{at~\dots} adverbials modify
activities. Cases where \qit{at~\dots} adverbials modify states or
points are covered by \pref{hpsg:100}.) \pref{hpsg:110} is the sign
for the reading where the situation of the culminating activity
\emph{starts} at the adverbial's time. The {\feat head} of
\pref{hpsg:109} is shown in \pref{hpsg:110}.
\begin{examples}
\item
\avmoptions{active}
\begin{avm}
[\avmspan{phon \; \<\fval at\>} \\
 ss|loc|cat & [head   & {\rm \pref{hpsg:110}} \\
               subj   & \<\>  \\
               comps  & \<\feat np$\,_{min\_per@1}$\> \\
               aspect & point]\\ 
 \avmspan{ss|cont \, $\at\[@1, \lbegin\[@2\]\]$}]
\end{avm}
\label{hpsg:109}
\item 
\avmoptions{active}
\begin{avm}
\osort{prep}{
[mod|loc|cat|aspect & culmact \\
 \avmspan{mod \; \feat s[vform {\fval fin}]:@2 $\lor$ 
                 \feat vp[vform {\fval psp}]:@2}]}
\end{avm}
\label{hpsg:110}
\end{examples}
Along with \pref{hpsg:2}, \pref{hpsg:109} causes \pref{hpsg:107} to be
mapped to \pref{hpsg:111}. \pref{hpsg:111} requires the period that
covers the entire inspection (from start to completion) to be located
in the past, and its start-point to fall within a 5:00pm-minute.
\begin{examples}
\item $\partop[\mbox{\textit{5:00pm}}^g, f^v] \land 
       \at[f^v, \lbegin[$ 
      $\past[e^v, \culm[inspecting(ep^v, ja, ba737)]]]]$
  \label{hpsg:111}
\end{examples}
The sign for the reading where the situation reaches its completion at
the adverbial's time is the same as \pref{hpsg:109}, except that it
introduces an \lend rather than a \lbegin operator. That sign leads to
a second formula for \pref{hpsg:107}, that contains an \lend instead
of a \lbegin. That formula corresponds to an unlikely reading in the
case of \pref{hpsg:107}, but it would capture the preferred reading in
\pref{hpsg:108}. A complete \nldb would typically paraphrase both formulae, and
ask the user to select the intended one (see \cite{DeRoeck1986} for
related discussion). Our prototype \nltdb simply attempts to generate a
separate answer for each formula.

We now explain briefly the reason for allowing temporal adverbials to
modify not only finite sentences, but also past-participle verb
phrases. In \pref{top:34} (see section \ref{pres-and-perf} above), 
this allows \qit{on 1/1/95} to attach
either to \qit{had J.Adams inspected BA737} or to \qit{inspected
BA737}. That is, \pref{top:34} is taken to have two possible parses,
sketched in \pref{hpsg:113} and \pref{hpsg:112}. These give rise to
\pref{top:35} and \pref{top:36} respectively, that express the
possible readings of \pref{top:34}.
\begin{examples}
\item {[}Had J.A.\ [inspected BA737]] on 1/1/95? \label{hpsg:113}
\item Had J.A.\ [[inspected BA737] on 1/1/95]? \label{hpsg:112}
\end{examples}
Like \qit{at~\dots} adverbials, \qit{on~\dots} adverbials introduce
\at operators. The \past and \perf operators of \pref{top:35} and
\pref{top:36} are introduced
by the \qit{had} auxiliary, while the \culm operator is
introduced by the lexical entry of the past participle \qit{inspected}. 
In \pref{hpsg:112}, the adverbial attaches
to \qit{inspected BA737} before \qit{had}. Hence, in \pref{top:36} the
\at ends up within the \past and \perf. In \pref{hpsg:113}, in
contrast, the \qit{had} attaches to the verb phrase first. This causes
the \at in \pref{top:35} to have wider scope over the \past and \perf.

\subsection{Fronted temporal modifiers}

When temporal modifiers attach to sentences, we allow them to either
follow or precede the sentences they modify.
This admits both \pref{hpsg:200} and \pref{hpsg:201}. Both sentences
receive \pref{hpsg:201b}.  (An alternative approach would be to allow
temporal modifiers to participate in unbounded dependencies; see pp.\
176 -- 181 of \cite{Pollard2}.)
\begin{examples}
\item Tank 2 was empty on Monday. \label{hpsg:200}
\item On Monday tank 2 was empty. \label{hpsg:201}
\item $\partop[\mathit{monday}^g, m^v] \; \land$ 
      $\at[m^v, \past[e^v, empty(tk2)]]$ \label{hpsg:201b}
\end{examples}
When temporal modifiers attach to past-participle verb phrases,
we require the modifiers to follow the verb phrases, as in
\pref{hpsg:202}. This rules out unacceptable sentences like
\pref{hpsg:203}. 
\begin{examples}
\item BA737 had [[entered sector 2] at 5:00pm]. \label{hpsg:202}
\item \bad BA737 had at 5:00pm entered sector 2. \label{hpsg:203}
\end{examples}

Our approach causes \pref{hpsg:204} to receive only \pref{hpsg:205}
(where the adverbial specifies the reference time), because in
\pref{hpsg:204} \qit{at 5:00pm} can modify only \qit{BA737 had entered
sector 2}; it cannot modify just \qit{entered sector 2} because of the
intervening \qit{BA737 had}.
\begin{examples}
\item At 5:00pm BA737 had entered sector 2. \label{hpsg:204}
\item $\partop[\mbox{\textit{5:00pm}}^g, f^v] \land \at[f^v,$ 
      $\past[e1^v, \perf[e2^v, enter(ba737,s2)]]]$ \label{hpsg:205}
\end{examples}
In contrast, \pref{hpsg:206} receives both \pref{hpsg:205} and
\pref{hpsg:207}, because in that case the adverbial can modify either
\qit{BA737 had entered sector 2} or \qit{entered sector 2}. In
\pref{hpsg:207} the adverbial specifies directly the time of the entrance.
\begin{examples}
\item BA737 had entered sector 2 at 5:00pm. \label{hpsg:206}
\item $\partop[\mbox{\textit{5:00pm}}^g, f^v] \land \past[e1^v,$ 
      $\perf[e2^v, \at[f^v, enter(ba737, s2)]]]$ \label{hpsg:207}
\end{examples}
The fact that \pref{hpsg:204} does not receive \pref{hpsg:207} does
not seem to be a disadvantage, because in \pref{hpsg:204} the reading
where the adverbial specifies directly the time of the entrance seems
unlikely. 

There is, however, a complication with our approach: when a sentence
is modified by both a preceding and a trailing temporal modifier (as in
\pref{hpsg:208}), two parses are generated: one where the trailing
modifier attaches first to the sentence, and one where the preceding
modifier attaches first. In \pref{hpsg:208}, this generates two
semantically equivalent formulae, \pref{hpsg:209} and
\pref{hpsg:210}. Our framework currently provides no 
mechanism to eliminate one of the two.
\begin{examples}
\item On 1/7/95 BA737 was at gate 2 for two hours. \label{hpsg:208}
\item $\at[\mbox{\textit{1/7/95}}, \for[\mathit{hour}^c, 2,$ 
      $\past[e^v, location\_of(ba737, g2)]]]$ \label{hpsg:209}
\item $\for[\mathit{hour}^c, 2, \at[\mbox{\textit{1/7/95}},$ 
      $\past[e^v, location\_of(ba737, g2)]]]$ \label{hpsg:210}
\end{examples}

\subsection{Temporal subordinate clauses}

Like temporal adverbials, temporal subordinate clauses are treated as
modifiers of finite sentences or past-participle verb phrases. In
\pref{hpsg:300}, for example, \qit{while runway 2 was open} is taken
to modify \qit{did any flight circle}. As with temporal adverbials,
when the modifier attaches to a sentence, it is allowed to either
precede or follow the sentence. This admits both \pref{hpsg:300} and
\pref{hpsg:301}.
\begin{examples}
\item Did any flight circle while runway 2 was open?
\label{hpsg:300}
\item While runway 2 was open, did any flight circle? \label{hpsg:301}
\end{examples}

Space does not permit a detailed presentation of our treatment of
temporal subordinate clauses. We note only that the words that
introduce the subordinate clauses (e.g.\ \qit{while}, \qit{before})
receive several signs in the lexicon, that are sensitive to the
aspectual classes of both the main and the subordinate clause. There
are, for example, two \qit{after} signs for the case where the main
clause is a point (e.g.\ \qit{which flights departed} in the airport
domain) and the \qit{after~\dots} clause is a state (e.g.\
\qit{system 32 was in operation}). The first sign introduces an \at
operator, whose first argument is the \topl representation of the
subordinate clause, and the second argument is the \topl
representation of the main clause. The second sign is similar, but it
also introduces a \lbegin operator in front of the formula of the
subordinate clause. These two signs cause \pref{hpsg:302} to receive
\pref{hpsg:303} and \pref{hpsg:304}. 
\begin{examples}
\item Which flights departed after system 32 was in operation? \label{hpsg:302}
\item $?f^v \; flight(f^v) \; \land$ 
      \hspace*{1.5mm}$\after[\!\!\!\begin{array}[t]{l}
              \past[e1^v, in\_oper(sys32)],
              \past[e2^v, depart(f^v)]]
             \end{array}$ \label{hpsg:303}
\item $?f^v \; flight(f^v) \; \land$ 
      \hspace*{1.5mm}$\after[\!\!\!\begin{array}[t]{l}
              \lbegin[\past[e1^v, in\_oper(sys32)]],
              \past[e2^v, depart(f^v)]]
             \end{array}$ \label{hpsg:304}
\end{examples}
\pref{hpsg:302} is ambiguous between a reading where the departures
must have happened after the \emph{beginning} of a maximal period
where system 32 was in operation, and a reading where the departures
must have happened after the \emph{end} of such a maximal
period. The two readings are captured by \pref{hpsg:303} and
\pref{hpsg:304} respectively. 

\subsection{Nominals and temporal reference} \label{nominals_time}

Following a distinction introduced in \cite{Pollard1}, we use
different signs for predicative and non-predicative uses of nouns
(e.g.\ \qit{president} is predicative in \pref{hpsg:400}, but
non-predicative in \pref{hpsg:401}). Predicative noun signs contribute
predicates ($president(w^v)$ in \pref{hpsg:400b}) that always end up
within the operators of the verb's tense (the \past in
\pref{hpsg:400b}). (There is a special sign for the non-auxiliary
\qit{was} in \pref{hpsg:400}. This introduces a \past operator, and
inserts the predicate of \qit{president} into the second
argument-slot of the \past.) This arrangement captures the fact that
predicative nouns always refer to the time of the verb tense ($e^v$ in
\pref{hpsg:400b}, the event time that has to fall within 1991).
\begin{examples}
\item Who was the president in 1991? \label{hpsg:400}
\item $?w^v \; \at[y1991, \past[e^v, president(w^v)]]$ \label{hpsg:400b}
\item The president visited Rome in 1991. \label{hpsg:401}
\end{examples}
In contrast, non-predicative nouns can generally refer either to the
time of the verb tense, or to the speech time. As already discussed in
section \ref{ntense_op}, for example, the \qit{president} of
\pref{hpsg:401} may refer to the person who was the president at the
time of the visit, or to the present president. Non-predicative nouns
can actually refer to any contextually salient time (see
\cite{Dalrymple1988}, \cite{Enc1986}, and \cite{Hinrichs}), but
limiting these salient times to $st$ and the time of the verb tense is
a reasonable simplification at this stage of development of \nltdb{s},
since a more thorough treatment would require a complete system
of temporal anaphora.

To account for the possible meanings of non-predicative nouns,
non-predicative noun signs place the nouns' predicates within \ntense
operators (e.g.\ $\ntense[t^v, president(p^v)]$). These predicates
(along with the \ntense{s}) are inserted into the quantifier storage,
and at the end of the parsing they are added in front of the rest of
the formula, outside the operators of the verb tenses. This causes
\pref{hpsg:401} to receive \pref{hpsg:401b}. 
\begin{examples}
\item $\ntense[t^v, president(p^v)] \; \land$ 
      \hspace*{1.5mm}$\at[y1991, \past[e^v, visiting(p^v, rome)]]$
\label{hpsg:401b} 
\end{examples}
In the formulae that are generated at the end of the parsing, the
\ntense{s} require the nouns' predicates to hold at unspecified times
($t^v$ in \pref{hpsg:401b}). During a subsequent post-processing phase
(to be discussed in section \ref{post_process}), each \ntense generally
leads to multiple formulae (\pref{top:47} and \pref{top:48} in the
case of \pref{hpsg:401b}), where the first argument of the \ntense has
been replaced by $now^*$ or a variable that occurs as a first argument
of a \past or \perf operator. These formulae capture the readings
where the noun refers to the speech time or the time of the verb tense
respectively. (In past perfect sentences, the ``time of the verb
tense'' can be either the reference time or the time of the past
situation.) In present sentences, like \pref{hpsg:402} that is
initially mapped to \pref{hpsg:403}, this arrangement correctly
generates only one formula, where the first argument of the \ntense is
$now^*$ (current president; in this case there is no \past or \perf
operator).
\begin{examples}
\item The president is visiting Rome. \label{hpsg:402}
\item $\ntense[t^v, president(p^v)] \; \land$
      \hspace*{1.5mm}$\pres[visiting(p^v, rome)]$ \label{hpsg:403}
\end{examples}
This strategy has the disadvantage that it leads to a multiplicity of
formulae in past sentences with several non-predicative nouns. Our
prototype \nltdb uses a rather ad hoc mechanism to reduce drastically
the number of generated formulae, that is based on the observation
that with many nouns, satisfactory answers can be obtained by assuming
that they refer always to $st$, or always to the time of the verb
tense. We do not discuss this mechanism here (see
\cite{Androutsopoulos1996}).

\subsection{Interrogatives} \label{interrogatives}

The interrogatives \qit{who} and \qit{what} are treated syntactically
as noun phrases. For example, \pref{hpsg:504} receives the same
syntactic analysis as \pref{hpsg:505}. (We ignore punctuation.) The
only difference from ordinary noun phrases is that \qit{who} and
\qit{what} introduce interrogative quantifiers. \pref{hpsg:504} and
\pref{hpsg:505} receive \pref{hpsg:504b} and \pref{hpsg:505b} respectively.
\begin{examples}
\item Who was inspecting BA737? \label{hpsg:504}
\item $?w^v \; \past[inspecting(ep^v, w^v, ba737)]$ \label{hpsg:504b}
\item J.Adams was inspecting BA737. \label{hpsg:505}
\item $\past[inspecting(ep^v, ja, ba737)]$ \label{hpsg:505b}
\end{examples}
Similarly, the interrogative \qit{which} is treated in the same way as
other noun-phrase determiners (e.g.\ \qit{a} in \pref{hpsg:511}),
except that it introduces an interrogative quantifier. \pref{hpsg:510}
and \pref{hpsg:511} receive \pref{hpsg:510b} and \pref{hpsg:511b}
respectively.
\begin{examples}
\item Which tank is empty? \label{hpsg:510}
\item $?k^v \! \begin{array}[t]{l}
               \ntense[now^*, tank(k^v)] \; \land 
               \pres[empty(k^v)]
               \end{array}$  \label{hpsg:510b} 
\item A tank is empty. \label{hpsg:511}
\item $\ntense[now^*, tank(k^v)] \land \pres[empty(k^v)]$ \label{hpsg:511b}
\end{examples}

Our \hpsg version also supports questions with multiple interrogatives, like
\pref{hpsg:520} which is mapped to \pref{hpsg:521}. \pref{hpsg:520}
receives the same syntactic analysis as \pref{hpsg:505}.
\begin{examples}
\item Who was inspecting what. \label{hpsg:520}
\item $\!\!\!\begin{array}[t]{l}?w1^v \; ?w2^v \;
      \past[e^v, inspecting(ep^v, w1^v, w2^v)]
      \end{array}$ \label{hpsg:521}
\end{examples}
Questions like \pref{hpsg:517}, where the interrogative refers to the
object of the verb, are treated using \hpsg's unbounded-dependencies
mechanisms (see \cite{Androutsopoulos1996} and \cite{Pollard2} for
details). 
\begin{examples}
\item Which flight did J.Adams inspect? \label{hpsg:517}
\end{examples}

Finally, the \qit{when} of \pref{hpsg:523} is treated syntactically as
a temporal-adverbial modifier of finite sentences. That is,
\pref{hpsg:523} receives the same syntactic analysis as
\pref{hpsg:525}. 
\begin{examples}
\item When was tank 2 empty? \label{hpsg:523} 
\item On 1/7/95 was tank 2 empty? \label{hpsg:525}
\end{examples}
The \qit{when} of \pref{hpsg:523} introduces a $?_{mxl}$ quantifier
(see section \ref{top:interrogs}). In
the formula that is generated at the end of the parsing, the variable
of this quantifier does not occur elsewhere in the formula (e.g.\
\pref{hpsg:523} initially receives \pref{hpsg:524}). During the
post-processing phase (discussed in section
\ref{post_process} below), this variable is replaced by the first
argument of a \past or \perf operator introduced by the main
clause. In \pref{hpsg:524}, this replaces $t^v$ by $e^v$, leading to a
formula that asks for the past maximal intervals where tank 2 was
empty.
\begin{examples}
\item $?_{mxl} t^v \; \past[e^v, empty(tk2)]$ \label{hpsg:524}
\end{examples}

\subsection{Post-processing} \label{post_process}

Before being translated into \tsql, the \topl formulae that are
generated by the parsing undergo an additional post-processing
phase. This is a collection of minor transformations that cannot be
carried out easily during the parsing. Two of these transformations
have already been discussed in sections \ref{nominals_time} and
\ref{interrogatives}: replacing the variables of \ntense operators and
$?_{mxl}$ quantifiers with variables that occur as arguments of \past
or \perf operators.

The third post-processing transformation is needed when
\qit{for~\dots} duration adverbials combine with verb forms that
normally imply that a climax was reached. In these cases, the
\qit{for~\dots} adverbials (when felicitous) cancel the normal
implication that the climax was reached. \pref{hpsg:600}, for example,
carries no implication that BA737 arrived at gate 2 (cf.\
\pref{hpsg:601}). The problem is that the parsing of \pref{hpsg:600}
generates \pref{hpsg:602}, which requires the taxiing to have been
completed.
\begin{examples}
\item BA737 taxied to gate 2 for five minutes. \label{hpsg:600}
\item $\!\!\!\begin{array}[t]{l}\for[\mathit{minute}^c, 5, 
      \past[e^v, \culm[taxiing\_to(ep^v, ba737, g2)]]
      \end{array}$ \label{hpsg:602}
\item BA737 taxied to gate 2. \label{hpsg:601}
\end{examples}
Roughly speaking, to solve this problem during the post-processing any
\culm operator that is within the scope of a \for operator is
removed. In \pref{hpsg:602}, this results in a formula that no longer
requires the taxiing to have been completed.

\qit{In~\dots} duration adverbials also introduce \for operators
(e.g.\ parsing \pref{hpsg:603} generates \pref{hpsg:602}).
\for operators are actually marked with a flag that shows if the
operator was introduced by a \qit{for~\dots} or \qit{in~\dots}
adverbial, and the post-processing removes only \culm operators from
within \for operators introduced by \qit{for~\dots} adverbials. This
distinction is necessary because unlike \qit{for~\dots} adverbials,
\qit{in~\dots} duration adverbials do not cancel the implication that the
climax was reached (\pref{hpsg:602} is the correct representation of
\pref{hpsg:603}).
\begin{examples}
\item BA737 taxied to gate 2 in five minutes. \label{hpsg:603}
\end{examples}

Admittedly the post-processing compromises the compositionality
of our English to \topl mapping. This, however, does not seem to have
any practical disadvantages. 

\subsection{The linguistic coverage} \label{linguistic_coverage}

We conclude the discussion of our English to \topl mapping with a list
of temporal linguistic phenomena that our framework can or cannot
handle. There is a wealth of temporal mechanisms in English (and most
natural languages), and it would have been impossible to consider all
of them in the time that we had available. Instead, we targetted a
carefully selected set of temporal mechanisms, which is both
non-trivial and representative of many types of temporal phenomena.
Of course, perfect grammatical/semantic descriptions are not
achievable, even within non-computational projects, so we are well
aware that our proposed analyses are no more than useful
simplifications.  Nevertheless, we believe that our framework can
serve as a basis for further work that will extend its coverage of
temporal phenomena (see also section \ref{further_work}).

\paragraph{Verb tenses:} We support six tenses: simple
present, simple past, present continuous, past continuous, present
perfect (treated as simple past), and past perfect. We have not
considered the past perfect continuous, future tenses, or futurate
readings of present and past tenses (e.g.\ the futurate reading of the
simple present -- see section \ref{aspect-criteria}). We do not
expect adding more tenses to be particularly difficult.

\paragraph{Temporal verbs:} Among special temporal verbs, we support
\qit{to start}, \qit{to begin}, \qit{to stop}, and \qit{to finish}. We
have not considered other special temporal verbs, like \qit{to happen}
or \qit{to follow}.

\paragraph{Temporal nouns:} Our framework supports nouns like
\qit{year}, \qit{month}, \qit{day}, etc.\ (and proper names like
\qit{Monday}, \qit{January}, \qit{1/5/92}). Nouns referring to more
abstract temporal entities (e.g.\ \qit{period}, \qit{event}), and
nouns that introduce situations (e.g.\ \qit{the construction of bridge
2}) are not supported. 

\paragraph{Temporal adjectives:} No temporal adjectives (e.g.\
\qit{first}, \qit{earliest}) are supported, with the exception of
\qit{current}, that we use to restrict the times to which some nouns
can refer (e.g.\ \qit{current president}; see section
\ref{nominals_time}).

\paragraph{Temporal adverbials:} Our framework supports temporal location
adverbials introduced by \qit{at}, \qit{on}, \qit{in}, \qit{before},
and \qit{after} (e.g.\ \qit{at 5:00pm}, \qit{on Monday}, \qit{in
January}, \qit{before 5:00pm}, \qit{after 1/1/95}), as well as
\qit{yesterday}, \qit{today}, and duration adverbials introduced by \qit{for}
and \qit{in} (\qit{for two days}, \qit{in two hours}). Other
adverbials that specify boundaries (e.g.\ \qit{since 5:00pm},
\qit{until 1/1/95}) should be easy to add. Frequency adverbials (e.g.\
\qit{twice}, \qit{daily}) seem more difficult to support. 

\paragraph{Temporal clauses:} Only temporal subordinate clauses introduced by
\qit{while}, \qit{before}, and \qit{after} are currently
handled. Adding support for other boundary-specifying clauses (e.g.\
\qit{since J.Adams became manager}, \qit{until tank 2 is empty})
should not be difficult. \qit{When~\dots} clauses are known to be the
most difficult to support (see \cite{Moens2} and \cite{Ritchie} for
related discussion).

\paragraph{Temporal anaphora:} Among temporal anaphoric phenomena, only the
anaphoric nature of nominals like \qit{the president} is supported
(section \ref{nominals_time}). As discussed in section
\ref{temp_advs}, our framework currently provides no mechanism to
determine the particular times to which adverbials like \qit{at
5:00pm} or verb tenses refer.


\section{The TSQL2 language} \label{TSQL2}

We now turn to \tsql, the database language into which we translate
the \topl formulae that the linguistic processing generates. \tsql
\cite{TSQL2book} is a temporal extension of \sqll \cite{Melton1993},
the \emph{de facto} standard database language. 
Traditional database languages such as
\sqll can and have been used to handle temporal
information, leading some researchers to question the need for
special temporal support in database languages (see \cite{Davies1995}
for related discussion). The lack of special temporal support in these
languages, however, makes it difficult to manipulate time-dependent
information. \tsql, for example, adds to \sqll a period datatype, and
keywords to compute the intersection of two periods, to check if two
periods overlap, etc. Without these, the rules that translate from
\topl to database language (to be discussed in section
\ref{TOP_to_TSQL2}) would be much more difficult to formulate, and
much harder to comprehend.

We also note that until recently there was little consensus among
temporal database researchers on how temporal support should be added
to \sqll (or other database languages), with every researcher
essentially adopting his/her own temporal database language. Perhaps
as a result of this, only a very few database management systems (\dbms{s})
that support these languages have been implemented (mostly early
prototypes; see \cite{Boehlen1995c}). In contrast, \tsql was designed
by a committee comprising most leading temporal database researchers,
and hence it has a much better chance of being supported by (or at
least influencing) forthcoming temporal database systems. 
As already mentioned in section \ref{introduction}, a prototype
database system (called \timedb) that supports a version of \tsql
has already appeared.

Like \sqll, \tsql is designed
to be used with the relational database model, where information is
stored in \emph{relations}, intuitively tables consisting of rows
(called \emph{tuples}) and columns (\emph{attributes}). As an example
of a \tsql relation, $tank\_contents$ shows the contents of various
tanks over time.

\adbtable{3}{|l|l||l|}{$tank\_contents$}
{$tank$  & $content$ & }
{$tank1$ & $oil$     & $\{[1/1/95, \; 31/5/95],$ \\
         &           & $\;\; [1/9/95, \; 1/5/96]\}$ \\
 $tank1$ & $empty$   & $\{[1/6/95, \; 31/8/95]\}$ \\
 $tank2$ & $water$   & $\{[1/1/95, \; 31/5/95],$ \\
         &           & $\;\; [1/8/95, \; 31/8/95]\}$ \\
 $tank2$ & $empty$   & $\{[1/6/95, \; 31/7/95],$ \\
         &           & $\;\; [1/9/95, \; 1/5/96]\}$ \\
 $tank3$ & $foam$    & $\{[1/1/95, \; 1/5/96]\}$
}
$tank\_contents$ has two \emph{explicit} attributes ($tank$ and
$content$), and an unnamed \emph{time-stamping} attribute that shows
when the information of each tuple was true (e.g.\ tank1 contained oil
from 1/1/95 to 31/5/95, and from 1/9/95 to 1/5/96). We use a double
line to separate the time-stamping attribute from the explicit
ones. $tank\_contents$ is in a normalised form where there is no
pair of tuples with the same explicit attribute values, and the
time-stamps (the values of the time-stamping attribute) are sets of maximal
periods. For simplicity, we assume in this example that time-stamps
are specified at the granularity of days. In the airport domain, the
time-stamps are actually specified at the granularity of minutes
(e.g.\ the last time-stamp of $tank\_contents$ above would be
something like
$\{[\mbox{8:32}am \; 1/1/95, \; \mbox{4:16}pm \; 1/5/96]\}$).

Information can be retrieved from the database using \tsql
\sql{SELECT} statements (these are temporally enhanced forms of the
corresponding \sqll statements). For example, \pref{tsql:1} generates
the relation in \pref{tsql:2}, that shows the tanks that were empty at
any time during July 1995.
\begin{examples}
\item \sql{\begin{tabular}[t]{l}
           SELECT SNAPSHOT t1.tank\\
           FROM tank\_contents AS t1 \\
           WHERE t1.content = 'empty' \\
           \ \   AND VALID(t1) OVERLAPS  PERIOD '[1/7/95 - 31/7/95]'
           \end{tabular}}
\label{tsql:1}
\item 
\dbtable{1}{|l|}{result of \pref{tsql:1}}
{$tank$ }
{$tank1$ \\
 $tank2$}
\label{tsql:2}
\end{examples}
The \sql{t1} in the \sql{FROM} clause of \pref{tsql:1} is a
\emph{correlation name}. It can be thought of as a tuple-variable that
ranges over the tuples of $tank\_contents$. The \sql{WHERE} clause
allows \sql{t1} to range over only tuples whose $content$ is $empty$,
and whose time-stamp overlaps the period from 1/7/95 to 31/7/95
(\sql{VALID(t1)} refers to the time-stamp of the
\sql{t1}-tuple). Finally, the \sql{SELECT} clause specifies that the
result should be a \emph{snapshot relation} (a relation with no
time-stamping attribute), whose only attribute contains the $tank$
values of the \sql{t1}-tuples.

As a further example, \pref{tsql:3} generates \pref{tsql:4}, where
each time-stamp is a single maximal period where the corresponding
tank was empty.
\begin{examples}
\item \sql{\begin{tabular}[t]{l}
           SELECT t1.tank\\
           VALID VALID(t1)\\
           FROM tank\_contents(PERIOD) AS t1 \\
           WHERE t1.content = 'empty'
           \end{tabular}}
\label{tsql:3}
\item \label{tsql:4}
\dbtable{2}{|l||l|}{result of \pref{tsql:3}}
{$tank$  &  }
{$tank1$ & $[1/6/95, \; 31/8/95]$ \\
 $tank2$ & $[1/6/95, \; 31/7/95]$ \\
 $tank2$ & $[1/9/95, \; 1/5/96]$
}
\end{examples}
The \sql{(PERIOD)} in the \sql{FROM} clause of \pref{tsql:3} generates
an intermediate relation that is the same as $tank\_contents$, except
that there is a separate tuple for each period in the time-stamps of
$tank\_contents$ (a tuple for $tank1$, $oil$, and $[1/1/95, \; 31/5/95]$,
a tuple for $tank1$, $oil$, and $[1/9/95, \; 1/5/96]$, etc.). The \sql{t1} of
\pref{tsql:3} ranges over tuples of the intermediate relation whose
$content$ is $empty$. The \sql{SELECT} and \sql{VALID} clauses specify
that the resulting relation should have one explicit and one
time-stamping attribute, and that the values of these attributes
should be the $tank$ values and time-stamps of the \sql{t1}-tuples.

To simplify the mapping from \topl to \tsql (to be discussed in
section \ref{TOP_to_TSQL2}), and to bypass some obscure points in the
definition of \tsql, we had to introduce some modifications in
\tsql.\footnote{Our work also revealed a number of possible improvements to
\tsql. These were reported in \cite{Androutsopoulos1995b}.} These are
relatively few and well-documented (in \cite{Androutsopoulos1996}), to
the extent that we are confident that they do not undermine the
research value of our \topl to \tsql mapping. For example, we assume
that attributes are ordered, and we refer to attributes by number
rather than by name (e.g.\ we write \pref{tsql:3b} instead of
\pref{tsql:3}).
\begin{examples}
\item \sql{\begin{tabular}[t]{l}
           SELECT t1.1\\
           VALID VALID(t1)\\
           FROM tank\_contents(PERIOD) AS t1 \\
           WHERE t1.2 = 'empty'
           \end{tabular}}
\label{tsql:3b}
\end{examples}
The most significant of our modifications is the addition of a
\sql{(SUBPERIOD)} keyword. This is intended to be used with relations
like \pref{tsql:4}, where each time-stamp is a single maximal
period. When applied to \pref{tsql:4}, \sql{(SUBPERIOD)} generates
\pref{tsql:4b}, that apart from the tuples of \pref{tsql:4} contains
tuples for all the subperiods of the original time-stamps. The use of
\sql{(SUBPERIOD)} is illustrated in \pref{tsql:3bb}, where
``\dots\pref{tsql:3b}\dots'' stands for the statement of
\pref{tsql:3b} (the current versions of \sqll and \tsql allow embedded
\sql{SELECT} statements in the \sql{FROM} clause). \pref{tsql:3bb}
generates a relation that contains all the tuples of \pref{tsql:4b}
whose time-stamp is a subperiod of $[4/6/95, \; 27/8/95]$.
\begin{examples}
\item \sql{\begin{tabular}[t]{l}
           SELECT t2.1 \\
           VALID VALID(t2) \\
           FROM (\mbox{\dots\pref{tsql:3b}\dots})(SUBPERIOD) AS t2 \\
           WHERE PERIOD '[4/6/95 - 27/8/95]' CONTAINS VALID(t2) 
           \end{tabular}}
\label{tsql:3bb}
\item \label{tsql:4b}
\dbtable{2}{|l||l|}{result of \pref{tsql:3b}}
{$tank$  &  }
{$tank1$ & $[1/6/95, \; 31/8/95]$ \\
 $tank1$ & $[3/6/95, \; 30/8/95]$ \\
 $tank1$ & $[5/6/95, \; 26/8/95]$ \\
 $tank1$ & $[1/7/95, \; 1/7/95]$ \\
 \dots   & \dots \\
 $tank2$ & $[1/6/95, \; 31/7/95]$ \\
 $tank2$ & $[4/6/95, \; 19/7/95]$ \\
 \dots   & \dots \\
 $tank2$ & $[1/9/95, \; 1/5/96]$ \\
 $tank2$ & $[3/9/95, \; 1/2/96]$ \\
 \dots   & \dots 
}
\end{examples}
\par\noindent
It is important that an implementation of the \sql{(SUBPERIOD)} construct
does not involve explicitly computing and storing this entire relation,
as this would involve vast amounts of space. The regular structure of
the relation, and the very fact that it is computable from a formula,
allows a database system to use a virtual relation, which behaves as
if it contained all the many tuples of the entire relation, without
the need for explicit representation of each one.


\section{From TOP to TSQL2} \label{TOP_to_TSQL2} 

We now discuss how \topl formulae are translated into \tsql. 
Although a method to translate from a form of temporal logic to
\tsql has already been proposed \cite{Boehlen1996}, that method is not
applicable to \nltdb{s}, because it is not coupled to a systematic
mapping from natural language to logic, and because the nature of the
logic of \cite{Boehlen1996} makes devising such a mapping very
difficult (this is discussed further in \cite{Androutsopoulos1996}). 

\subsection{Primitive TOP expressions} \label{primitive-maps}

One central issue for any kind of \nldb is that of portability.
Customising an \nldb or \nltdb to handle a new or different domain is
bound to involve some human effort in defining how natural language
expressions or lexical entries relate to symbolic structures in the
database and/or expressions in the database query language. A
well-designed system will limit the range of structures that have to
be altered, and make it very clear what items must be edited and in
what way.  We have attempted to do this by specifying a well-defined
set of functions (the $h'$ functions of figure \ref{simple-arch-fig})
which form this particular interface.

The person who configures the \nltdb for a new database has to specify
how primitive \topl expressions like constants and predicate functors
relate to \tsql expressions. In the case of \topl constants, the
configurer has to specify a function $\hconsp$ that maps each constant
to a \tsql literal that represents the same entity in the world. For
example, the \topl constant $tk2$ (used, for example, in
\pref{hpsg:524}) represents the same tank as the attribute value
$tank2$ in $tank\_contents$, and this attribute value is written in
\tsql as \sql{'tank2'}; i.e.\ $\hconsp(tk2) =$
\sql{'tank2'}. Similarly, we use the \topl constant $y1991$ (e.g.\
in \pref{hpsg:401b}) to refer to the time-period that is written in
\tsql as \sql{PERIOD '[1/1/91 - 31/12/91]'}; i.e.\ $\hconsp(y1991) =$
\sql{PERIOD '[1/1/91 - 31/12/91]'}.

In the case of \topl predicate functors (e.g.\ $empty$ in \pref{hpsg:524}),
the configurer has to specify a function $\hpfunsp$ that maps each
functor to a \tsql \sql{SELECT} statement.\footnote{
In \cite{Androutsopoulos1996} \hpfunsp has an additional argument that 
corresponds to the arity of the predicate. For simplicity, we ignore this
argument here.}
That statement must
generate a relation of the same form as \pref{tsql:4} that shows for each
combination of predicate arguments, the maximal periods where the
situation of the predicate is true (i.e.\ $\hpfunsp(empty)$ would be
the statement \pref{tsql:3} that generates
\pref{tsql:4}).\footnote{This assumes that each predicate can be
mapped directly to a relation computed from the contents of the
database. There are cases (e.g.\ the ``doctor-on-board'' problem
\cite{Rayner93}) where this is not possible, and an intermediate
reasoning layer is needed. See \cite{Androutsopoulos1996} for
discussion on how this reasoning layer would fit into our framework.} With
predicate functors introduced by culminating activity verbs
(e.g.\ $inspecting$ in \pref{hpsg:521}), the configurer specifies an
additional \sql{SELECT} statement. This generates a relation that
shows for each combination of predicate arguments, whether or not the
situation of the predicate has reached its climax. We do not discuss
this here (see \cite{Androutsopoulos1996}). 

To some extent, the specification of $h'$ functions can be supported
by standard templates for the configurer to use, or by adopting rewrite
rules that specify the values of the  $h'$ functions for whole classes of \topl
expressions. In a particular domain, for example, it may be appropriate 
to use a rewrite rule stating that \topl constants containing underscores
are mapped to multi-word \tsql strings by replacing underscores with
spaces and capitalising the first letter of each word in the \tsql string
(this sets \hconsp(united\_kingdom) = \sql{'United Kingdom'}). Rewrite 
rules of this kind can be easily incorporated into our prototype \nltdb, 
where the $h'$ functions are implemented as Prolog rules.

\subsection{The translation rules}

The \topl formulae are translated into \tsql by a set of translation
rules of the form:
\[ trans(\phi, \lambda) = \Sigma \]
where $\phi$ is the formula to be translated, $\lambda$ is a \tsql
expression that intuitively represents what was the localisation time
when $\phi$ was encountered (this will become clearer below), and
$\Sigma$ is the \tsql translation of $\phi$ ($\Sigma$ is always a 
\sql{SELECT} statement). There are: (i) basic (non-recursive)
translation rules for the cases where $\phi$ is a atomic formula or a
formula of the form $\culm[\phi']$, and (ii) recursive translation
rules for more complex formulae, that translate $\phi$ by calling
other translation rules to translate subformulae of $\phi$. 

For example, ignoring some details, the translation rule for the case
where $\phi$ is an atomic formula $\pi(\tau_1, \dots, \tau_n)$ is the
following:
\medskip
\\
$trans(\pi(\tau_1, \dots, \tau_n), \lambda) =$\\
       \select{SELECT $\alpha.1$, $\alpha.2$, \dots, $\alpha.n$ \\
               VALID VALID($\alpha$) \\
               FROM ($\hpfunsp(\pi)$)(SUBPERIOD) AS $\alpha$ \\
               WHERE \dots \\
               \ \ AND \dots \\
               \ \ \vdots \\
               \ \ AND \dots \\
               \ \ AND $\lambda$ CONTAINS VALID($\alpha$)} 
\medskip
\\ 
Whenever the translation rule is used, $\alpha$ is a new
correlation name (section \ref{TSQL2}). The ``\dots'' in the
\sql{WHERE} clause are all the strings in $S_1 \union S_2$:
\begin{eqnarray*}
S_1 &=&
\{\mbox{``}\alpha.i = \hconsp(\tau_i)\mbox{''} \mid
  i \in \{1,2,3,\dots,n\} \\
    && \mbox{ and }\tau_i \mbox{ is a \topl constant}\} \\
S_2 &=& 
\{\mbox{``}\alpha.i = \alpha.j\mbox{''} \mid
  i,j \in \{1,2,3,\dots,n\}, \\
    && \mbox{ } \tau_i = \tau_j, \mbox{ and }
        \tau_i, \tau_j \mbox{ are \topl variables}\}
\end{eqnarray*}

\noindent Informally, $S_2$ contains restrictions that are needed when
a variable appears at more than one predicate-argument positions. The
$S_2$ restrictions ensure that the attributes of the resulting
relation that correspond to these argument positions have the same
values in each tuple.

Let us assume, for example, that we use the rule above to translate
the formula $empty(tk^v)$, and that $\lambda$ is set to the
$\lambda_1$ of \pref{trans:3x}, where $\chi$ is the \tsql name of the
finest available granularity (the level of \emph{chronons} in \tsql
terminology). The \sql{TIMESTAMP 'now' - INTERVAL '1' $\chi$} below
refers to the time-point immediately before $st$.
\begin{examples} 
\item $\lambda_1 =$ \sql{INTERSECT(}\\
      \sql{\begin{tabular}[t]{l}
           PERIOD($\!\!\!$\begin{tabular}[t]{l}
                  TIMESTAMP 'beginning', \\
                  TIMESTAMP 'forever),
                  \end{tabular} \\
           PERIOD($\!\!\!$\begin{tabular}[t]{l}
                  TIMESTAMP 'beginning', \\
                  TIMESTAMP 'now' - INTERVAL '1' $\chi$))
                  \end{tabular}
          \end{tabular}}
\label{trans:3x}
\end{examples}
$\lambda_1$ is the \tsql representation of $[t_{first}, \; t_{last}]
\intersect [t_{first}, \; st)$, where $t_{first}$ and $t_{last}$ are
the earliest and latest time-points respectively (like \topl, \tsql
assumes that time is discrete and bounded). $[t_{first}, \; t_{last}]
\intersect [t_{first}, \; st)$ is, of course, equal to $[t_{first}, \;
st)$. We use $\lambda_1$ here to make this example easier to integrate
with later examples.  The translation rule maps $empty(tk^v)$ to
\pref{trans:3}, where $\hpfunsp(empty)$ is \pref{tsql:3}. Assuming
that 1/5/96 is in the past, \pref{trans:3} generates \pref{tsql:4b},
that shows for all the possible values of $tk^v$, the past event-time
periods where $empty(tk^v)$ is true that fall within $lt$ (the period
of $\lambda_1$). (The \sql{WHERE} clause of \pref{trans:3} has no
effect in this case, because $lt = [t_{first}, \; st)$, and all the
time-stamps of the \sql{t2}-tuples are subperiods of $lt$.)
\begin{examples}
\item \select{SELECT t2.1 \\
              VALID VALID(t2) \\
              FROM ($\hpfunsp(empty)$)(SUBPERIOD) AS t2 \\
              WHERE $\lambda_1$ CONTAINS VALID(t2)}
\label{trans:3}
\end{examples}
In this particular example, $S_1$ and $S_2$ are empty. If the
predicate contained constants as arguments (e.g.\ $empty(tk2)$), $S_1$
would introduce restrictions in the \sql{WHERE} clause that would
cause the resulting relation to contain information only about the
world entities of those constants. In $empty(tk2)$, for example,
\pref{trans:3} would contain the additional restriction \sql{t2.1 =
'tank2'} in its \sql{WHERE} clause, which would remove from the result
tuples that do not refer to tank 2. 

Let us now consider the recursive translation rule for the case where
$\phi$ has the form $\past[\beta, \phi']$ ($\beta$ is a \topl
variable, and $\phi'$ a formula):
\medskip \\
\noindent $trans(\past[\beta, \phi'], \lambda) =$\\
       \select{SELECT VALID($\alpha$), 
                      $\alpha$.1, $\alpha$.2, \dots, $\alpha$.$n$ \\
               VALID VALID($\alpha$) \\
               FROM ($trans(\phi', \lambda')$) AS $\alpha$} \\
\\
Every time the translation rule is used, $\alpha$ is a new correlation
name. $n$ is the number of explicit attributes of the relation
generated by $trans(\phi', \lambda')$, and $\lambda'$ is the \tsql
expression:
\par\noindent
\sql{INTERSECT(}$\lambda$,
\sql{PERIOD(TIMESTAMP 'beginning',TIMESTAMP 'now' - INTERVAL '1' $\chi$))}
\smallskip
\par\noindent
The translation rule for $\past[\beta, \phi']$ narrows the
period of $\lambda$ (intuitively $lt$) to its intersection with
$[t_{first}, st)$ (this is in accordance with the semantics of the
\past operator; see section \ref{TOP_basics}).

Let us assume, for example, that the rule above is used to translate
$\past[e^v, empty(tk^v)]$, and that $\lambda$ has been set to the
$\lambda_2$ of \pref{trans:10} (that specifies a period that covers
the entire time-axis). $\past[e^v, empty(tk^v)]$ would be mapped to
\pref{trans:11}, where $trans(empty(tk^v), \lambda_1)$ is
\pref{trans:3}.
\begin{examples}
\item $\lambda_2 =$ \sql{PERIOD(}$\!\!\!$\begin{tabular}[t]{l}
                    \sql{TIMESTAMP 'beginning',}
                    \sql{TIMESTAMP 'forever')}
                    \end{tabular}
   \label{trans:10}
\item \select{SELECT VALID(t3), t3.1 \\
              VALID VALID(t3) \\
              FROM ($trans(empty(tk^v), \lambda_1)$) AS t3}
   \label{trans:11}
\end{examples}
The translation rules for formulae with no interrogative quantifiers
(e.g.\ $\past[\beta, \phi']$) generate \sql{SELECT} statements that
return relations containing one explicit attribute for each
predicate-argument and each variable that occurs as argument of a
temporal operator. \pref{trans:11} returns \pref{trans:12}, where the
first and second explicit attributes correspond to $e^v$ and $tk^v$
respectively, and the time-stamps represent event-time periods where
$\past[e^v, empty(tk^v)]$ holds, for the various combinations of $e^v$
and $tk^v$ values.
\begin{examples}
\item \label{trans:12}
\dbtableb{|l|l||l|}
{\dots & $tank1$ & $[1/6/95, \; 31/8/95]$ \\
 \dots & $tank1$ & $[3/6/95, \; 30/8/95]$ \\
 \dots & $tank1$ & $[5/6/95, \; 26/8/95]$ \\
 \dots & $tank1$ & $[1/7/95, \; 1/7/95]$ \\
 \dots & \dots   & \dots \\
 \dots & $tank2$ & $[1/6/95, \; 31/7/95]$ \\
 \dots & $tank2$ & $[4/6/95, \; 19/7/95]$ \\
 \dots & \dots   & \dots \\
 \dots & $tank2$ & $[1/9/95, \; 1/5/96]$ \\
 \dots & $tank2$ & $[3/9/95, \; 1/2/96]$ \\
 \dots & \dots   & \dots
}
\end{examples}

The \sql{SELECT} clause of \pref{trans:11} specifies that
\pref{trans:12} must have two explicit attributes, and that in each
tuple the first explicit attribute (the one for $e^v$) must have the
same value as the time-stamp. (To save space, we do not show the
values of the first explicit attribute in \pref{trans:12}.) This
reflects the fact that in $\past[e^v, empty(tk^v)]$ the semantics of
\past requires the value of $e^v$ to be the event time of
$empty(tk^v)$ (section \ref{top:interrogs}).

When translating from \topl to \tsql, $\lambda$ is initially set to
the $\lambda_2$ of \pref{trans:10}. (This reflects the fact that
$lt$ initially covers the whole time-axis -- section
\ref{TOP_basics}.) For example, the \tsql translation of \pref{trans:13}
(\qit{Was anything empty?} (at any past time)) is $trans(\past[e^v,
empty(tk^v)], \lambda_2)$, i.e.\ \pref{trans:11}. In the case of yes/no
questions, the answer is affirmative if the generated \tsql code
returns a relation with at least one tuple, and negative otherwise. In
\pref{trans:13}, the generated \tsql code (i.e.\ \pref{trans:11})
returns \pref{trans:12}, and hence the answer would be affirmative. 
\begin{examples}
\item $\past[e^v, empty(tk^v)]$ \label{trans:13}
\end{examples}

As a final example of a translation rule, we show below the rule for
the case where $\phi$ has the form $?\beta_1 \; ?\beta_2 \; \dots ?\beta_k \;
\phi'$ ($\beta_1, \beta_2, \dots, \beta_k$ are \topl variables,
and $\phi'$ a formula). This type of formula is intended to reflect
the meaning of \emph{wh}-questions in English, with each $?\beta_i$
corresponding to a wh-word (\qit{what}, \qit{which}, etc.; see
sections \ref{top:interrogs} and \ref{interrogatives}). 
\medskip
\\
$trans(?\beta_1 \; ?\beta_2 \; \dots \;
?\beta_k \; \phi', \lambda) =$ \\
       \select{SELECT SNAPSHOT $\alpha.\omega_1$,
                      $\alpha.\omega_2$, \dots, $\alpha.\omega_k$ \\
               FROM $trans(\phi', \lambda)$ AS $\alpha$} 
\medskip
\\
Whenever this rule is used, $\alpha$ is a new correlation
name. $\omega_1$, $\omega_2$,~\dots, $\omega_k$ are numbers referring
to the explicit attributes of the relation returned by $trans(\phi',
\lambda)$. The $\omega_1$-th, $\omega_2$-th, \dots, $\omega_k$-th
explicit attributes of that relation must correspond to $\beta_1$,
$\beta_2$, \dots, $\beta_k$ respectively. (This correspondance is
defined formally in \cite{Androutsopoulos1996}.)

For example, the \tsql translation of $\phi = ?tk^v \; \past[e^v,
empty(tk^v)]$ is $trans(\phi, \lambda_2)$, which according to the
translation rule above is \pref{trans:15}. (The $trans(\past[e^v,
empty(tk^v)], \lambda_2)$ in \pref{trans:15} is \pref{trans:11}.)
\begin{examples}
\item \select{SELECT SNAPSHOT t4.2 \\
              FROM ($trans(\past[e^v, empty(tk^v)], \lambda_2)$) AS t4}
   \label{trans:15}
\end{examples}
\pref{trans:15} generates a one-attribute snapshot relation that lists
all the tanks whose names appear in the second explicit attribute of
\pref{trans:12} (i.e.\ tanks that were empty in the past).

\subsection{Outline of the correctness proof}

We have proven formally (in \cite{Androutsopoulos1996}) the
correctness of the \topl to \tsql translation rules.  The full details
of this proof are highly formal and voluminous, and hence beyond the
scope of this paper. Here, we only attempt to provide an outline of
our approach.

We assume that a \topl expression and a \tsql query both refer to some
universe of world entities, relationships, etc.\ (much as in
first-order logic), including temporal entities (such as periods). The
denotation, in terms of this universe, of a \topl formula is provided
by the definition of \topl's semantics (see
\cite{Androutsopoulos1996}). The denotation of a \tsql expression is
given indirectly, in that we assume that there is an ``evaluation''
function which will map a \tsql query to some database constructs
(attribute values, relations, etc.), and the semantics of the database
indicates how such constructs are related to the universe of world
entities, relationships, periods, etc. The situation is roughly as in
figure \ref{translation_fig}. 

We have proven that the translation rules are correct, in the sense
that the denotation of a \topl formula (roughly speaking its answer)
as defined by the semantics of \topl is the same as the denotation
(answer) of the corresponding \tsql query when determined by way of
the ``evaluation'' function and the database semantics. In terms of
figure \ref{translation_fig} above, the same semantic content is
reached whether path 1 or path 2 is chosen.

\begin{figure}
\hrule
\begin{center}
\medskip
\includegraphics[scale=.6]{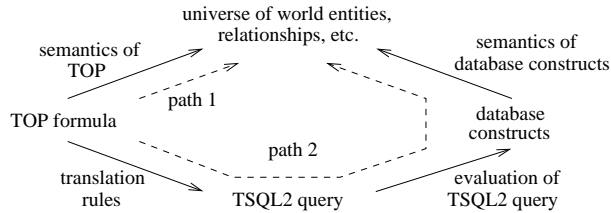}
\caption{From \topl to world entities}
\label{translation_fig}
\end{center}
\hrule
\end{figure}


\section{Comparison to previous work} \label{previous_work}

Although there has been a significant body of work on tense and aspect
theories, temporal logics, and temporal database, only a few
researchers have attempted to apply this work on \nldb{s}, most
notably Clifford \cite{Clifford3,Clifford,Clifford2}, De
et al.\ \cite{De,De2}, and Moens \cite{Moens,Moens3,Moens2}.
 
Clifford's work constitutes the most significant previous approach to
\nltdb{s}. Clifford defines formally a temporal version of the
relational database model, and a fragment of English that can be used
to query databases structured according to his model. Following the
Montague tradition \cite{Dowty}, Clifford employs an intensional
higher-order language (called \ils) to represent the semantics of the
English questions, and a temporally enhanced version of Montague's
\ptq grammar to map from English to \ils. Clifford's coverage of
English, however, is extremely narrow, and the semantics of the
English temporal mechanisms are oversimplified. For example, perfect
and continuous tenses are not supported, and no aspectual taxonomy
(section \ref{aspectual_classes}) is employed.  Clifford also defines
a version of relational algebra \cite{Ullman} for his database model
(this can be thought of as a theoretical database query language), and
sketches a method to translate from \ils to that algebra. The
description of this mapping, however, is very informal, and
with no proof of its correctness. There is also no indication that
Clifford's theory was ever used to implement an actual \nltdb.

De et al.\ describe a question-answering system that can handle a
fragment of English questions involving time. The ``temporal
database'' in this case is a rather {\it ad hoc} collection of facts
and inference rules, expressed in a kind of logic-programming
language. There is no clear intermediate meaning representation
language, and it is very difficult to distinguish the part of the
system that is responsible for the linguistic processing from the part
of the system that is responsible for retrieving information from the
database. De et al.\ consider this an advantage, but it clearly
sacrifices modularity and portability (cf.\  remarks 
in section \ref{primitive-maps} above). For example, it is very hard
to see which parts of the software would have to be modified if the
natural language processor were to be used with a commercial \dbms.
De et al.'s system does not seem to be based on any clear linguistic
analysis. There is also very little information in \cite{De} and
\cite{De2} on exactly which temporal linguistic mechanisms are
supported, and which semantics are assigned to these mechanisms.

Moens' work on temporal linguistic phenomena has been highly
influential in the area of tense and aspect theories. In the last part
of \cite{Moens} (see also \cite{Moens3}), Moens develops a simplistic
\nltdb. This has a very limited linguistic coverage, and is mainly
intended to illustrate Moens' tense and aspect theory, rather than to
constitute a detailed exploration of \nltdb{s}. As in the case of De
et al., Moens' ``database'' is a rather idiosyncratic collection of
facts and inference rules (expressed in Prolog). Moens' system uses a
subset of Prolog as its meaning representation language. English
questions are translated into expressions of this subset using a \dcg
grammar \cite{Pereira1980}, and there are Prolog rules that directly
evaluate the meaning representation language expressions against the
database. Moens provides essentially no information about the \dcg
grammar. The definition of the meaning representation language is also
very unclear (e.g.\ it is difficult to see exactly which Prolog
expressions are part of the representation language), and the
semantics of the language is defined very informally (by listing
Prolog code that evaluates some of the possible expressions of the
representation language).

We argue in \cite{Androutsopoulos1996} that all previous approaches to
\nltdb{s} suffer from one or more of the following: (i) they ignore
important English temporal mechanisms, or (ii) they assign to them
over-simplified semantics (both (i) and (ii) apply to Clifford and De et al.),
(iii) they lack clearly defined meaning representation languages (De et
al.\ and Moens), (iv) they do not provide complete descriptions of
the mappings from natural language to meaning representation language
(De et al., Moens), or (v) from meaning representation language to
database language (Clifford), (vi) they adopt idiosyncratic (Clifford)
and possibly also not well-defined database models or languages (De et
al., Moens), (vii) they do not demonstrate that their ideas are
implementable (Clifford).

Our framework avoids completely pitfalls (iii) -- (vi): our meaning
representation language (\topl) is completely and formally defined
(see appendix \ref{top-definitions}, and also \cite{Androutsopoulos1996}); the mapping from English to \topl is
based on a well-known and widely used grammar framework (\hpsg), and
all our modifications to it are well-documented (in
\cite{Androutsopoulos1996}); we have adopted a database language
(\tsql) and its underlying database model that the temporal database
community is considering as a basis for a standard; and we have developed a formally
defined (and provably correct) method to translate from \topl to
\tsql.

Our proposal is open to criticism on (i), in the sense that several
important temporal linguistic mechanisms remain to be added to our
linguistic coverage. The temporal mechanisms that our framework
supports, however, are assigned sufficiently elaborate semantics, and
therefore our framework does not suffer from (ii). Finally, our
framework is partially open to criticism on point (vii), in the sense
that our prototype \nltdb has not been linked to an actual \dbms. We
are confident that our framework is sufficiently well-documented and
concretely founded, to the extent that future work will be able to
improve it with regard to points (i) and (vii).


\section{Conclusions and further work} \label{further_work} 

We have presented a theoretical framework for building \nltdb{s}, and
a prototype \nltdb that proves the implementability of that
framework. We have also argued that our framework constitutes an
improvement over previous work on \nltdb{s}. We discuss below some
possible extensions to our work. 

\paragraph{Extending the linguistic coverage:} We have already pointed
out the need to extend the linguistic coverage of our framework. Among
the temporal linguistic mechanisms that we have not considered
(section \ref{linguistic_coverage}), temporal anaphoric phenomena
(discussed briefly in section \ref{temp_advs}) seem to be particularly
interesting: several researchers have examined temporal
anaphora (e.g.\ \cite{Eberle1989}, \cite{Hinrichs1986},
\cite{Kamp1993}, \cite{Partee1984}, \cite{Webber1988}), and it would
be interesting to explore the applicability of their proposals to
\nltdb{s}. A Wizard of Oz experiment could also be performed to
determine which temporal phenomena most urgently need to be added to
the linguistic coverage of our framework, and to collect sample
questions that could be used as a test suite for \nltdb{s}
\cite{King1996}. (In a Wizard of Oz experiment, users interact through
terminals with a person that pretends to be a natural language
interface; see \cite{Diaper1986}.)

\paragraph{Cooperative responses:} In \cite{Androutsopoulos1996} we
reach the conclusion that \emph{cooperative responses}
\cite{Kaplan1982} are particularly important in \nltdb{s}, and that a
mechanism to generate such responses should be added to our framework.
For example, \pref{coop:2} is currently mapped to a \topl formula that
requires BA737 to have \emph{reached} gate 2 for the answer to be
affirmative. This causes a negative response to be generated if BA737
was taxiing to gate 2 but never reached it. While a simple negative
response is strictly speaking correct, it is hardly satisfactory in
this case. A more cooperative response like \pref{coop:4} is needed.
\begin{examples}
\item Did BA737 taxi to gate 2? \label{coop:2}
\item \sys{BA737 was taxiing to gate 2, but never reached it.} 
      \label{coop:4}
\end{examples}
As noted by \cite{Kaplan1982} and others, cooperative responses can
also be a useful enhancement to a \nldb's capabilities even where
the reply is positive:
\begin{examples}
\item Did BA737 taxi to gate 2? \label{coop:3}
\item \sys{BA737 taxied to gate 2 at 9:05am on 10/11/94} \label{coop:5} \\
\end{examples}

\vspace*{-5mm}
\paragraph{Linking to a DBMS:} We have already discussed the need to
link the prototype \nltdb to a \tsql \dbms (\timedb being currently
the only option we know of). 

\paragraph{Extending the prototype NLTDB:} As already noted in section
\ref{introduction}, several components would have to be added to the
prototype \nltdb, if it were to be used in real-life applications
(e.g.\ a lexical preprocessor, a module for paraphrasing the user's
questions, a module for quantifier-scoping, a reasoning component,
configuration tools, etc.; see \cite{Androutsopoulos1996} for
discussion on how these would fit into our existing architecture). As
most of these components are not directly sensitive to temporal
issues, it should be possible to implement them by borrowing existing
techniques from conventional (non-temporal) \nldb{s}.

\paragraph{Optimising the TSQL2 code:} Although provably correct, the
queries that the \topl to \tsql translator generates are often
verbose. It is usually easy for a \tsql programmer to see ways to
shorten significantly the generated \tsql code. Long database
language expressions can confuse \dbms{s} (especially prototype ones),
leading them to unacceptable performance.  A \tsql optimiser is needed
that will apply simplifications to the \tsql code of the \topl to
\tsql translator, before the \tsql queries are passed to the \dbms
(see \cite{Androutsopoulos1996} for related discussion). 


\newpage
\bibliographystyle{plain}
\bibliography{biblio}


\newpage
\appendix
\section{Appendix : The Top Language}\label{top-definitions}

\subsection{The Syntax of TOP}\label{top-syntax}

\subsubsection*{Notation}

\begin{tabular}{lll}
 \{  \} & : & Used to group elements in BNF. Not part of TOP language. \\
 $F^*$  & : & Zero or more repetitions of $F$. \\
 $F^+$  & : & One or more repetitions of $F$. \\
 $\mid$ & : & Separates different possible RHSs of BNF rule. \\
 $\langle n\rangle$ & : & Footnote -- indicates some further condition on 
                          rule.
\end{tabular}

\medskip

\noindent Terminal symbols are in lower case, usually with an initial
capital.  Nonterminals are in upper case.

\subsubsection*{Grammar}

The distinguished symbol is $\forms$.

\begin{verse}
$ \forms  \rightarrow  \ynforms  \ \ \mid\ \   \whforms $\\

$ \whforms \rightarrow  \whformsone \ \ \mid\ \  \whformstwo $\\

$ \whforms1  \rightarrow \{? \vars\}^{+} \ynforms \hfill\gnote{1} $\\

$ \whforms2  \rightarrow ?_{mxl} \vars \{? \vars\}^{*}  \ynforms  \hfill \gnote{2} $\\
$ \ynforms  \rightarrow  \aforms $\\
$\ \ \ \ \ \ \ \ \ \ \ \ \ \ \ \ \ \ \ \  \mid \ynforms  \wedge \ynforms $\\
$\ \ \ \ \ \ \ \ \ \ \ \ \ \ \ \ \ \ \ \  \mid \pres[\ynforms] $\\
$\ \ \ \ \ \ \ \ \ \ \ \ \ \ \ \ \ \ \ \  \mid \past[\vars, \ynforms]\ \ \ \ \ \ \ \ \ \     \hfill \gnote{3} $\\
$\ \ \ \ \ \ \ \ \ \ \ \ \ \ \ \ \ \ \ \  \mid \perf[\vars,\ynforms]\ \ \ \ \ \ \ \ \ \      \hfill \gnote{3} $\\
$\ \ \ \ \ \ \ \ \ \ \ \ \ \ \ \ \ \ \ \  \mid \at[\terms,\ynforms]\ \ \ \ \ \ \ \ \ \        \hfill \gnote{4} $\\
$\ \ \ \ \ \ \ \ \ \ \ \ \ \ \ \ \ \ \ \  \mid \at[\ynforms,\ynforms] $\\
$\ \ \ \ \ \ \ \ \ \ \ \ \ \ \ \ \ \ \ \  \mid \before[\terms,\ynforms]\ \ \ \ \ \ \ \ \ \    \hfill \gnote{4} $\\
$\ \ \ \ \ \ \ \ \ \ \ \ \ \ \ \ \ \ \ \  \mid \before[\ynforms, \ynforms] $\\
$\ \ \ \ \ \ \ \ \ \ \ \ \ \ \ \ \ \ \ \  \mid \after[\terms,\ynforms]\ \ \ \ \ \ \ \ \ \    \hfill \gnote{4} $\\
$\ \ \ \ \ \ \ \ \ \ \ \ \ \ \ \ \ \ \ \  \mid \after[\ynforms, \ynforms] $\\
$\ \ \ \ \ \ \ \ \ \ \ \ \ \ \ \ \ \ \ \  \mid \ntense[\vars, \ynforms]\ \ \ \ \ \ \ \ \ \   \hfill \gnote{3} $\\
$\ \ \ \ \ \ \ \ \ \ \ \ \ \ \ \ \ \ \ \  \mid \ntense[now*, \ynforms] $\\
$\ \ \ \ \ \ \ \ \ \ \ \ \ \ \ \ \ \ \ \  \mid \for[\cparts, VQTY, \ynforms] $\\
$\ \ \ \ \ \ \ \ \ \ \ \ \ \ \ \ \ \ \ \  \mid \fills[\ynforms] $\\
$\ \ \ \ \ \ \ \ \ \ \ \ \ \ \ \ \ \ \ \  \mid \lbegin[\ynforms] $\\
$\ \ \ \ \ \ \ \ \ \ \ \ \ \ \ \ \ \ \ \  \mid \lend[\ynforms] $\\
$\ \ \ \ \ \ \ \ \ \ \ \ \ \ \ \ \ \ \ \  \mid \culm[LITERAL] $\\

$ \aforms  \rightarrow  LITERAL  \ \ \mid\ \  \partop[\parts,\vars]   \ \ \mid\ \  \partop[\parts, \vars, VORD] $\\

$ LITERAL \rightarrow \pfuns(\{\terms ,\}^* \terms) $\\

$ \terms  \rightarrow \cons  \ \ \mid\ \  \vars $\\

$ \parts  \rightarrow  \cparts \ \ \mid\ \  \gparts $\\

$ VORD  \rightarrow 0 \ \ \mid\ \  -1 \ \ \mid\ \  -2 \ \ \mid\ \  \ldots $\\

$ VQTY  \rightarrow 1 \ \ \mid\ \  2 \ \ \mid\ \  3 \ \ \mid\ \  \ldots $\\

\end{verse}

\noindent {\sc \pfuns, \cparts, \gparts, \cons, \vars} are disjoint
open classes of terminal symbols.

\subsubsection*{Footnotes}

\begin{itemize}

\item[$\gnote{1}$] : Each $\vars$ terminal which appears in the $\{?
\vars\}^{+}$ expression must occur inside the $\ynforms$ expression.

\item[$\gnote{2}$] : Each $\vars$ terminal which appears in the $\{?
\vars\}^{+}$ expression must occur inside the $\ynforms$ expression, and the
$\vars$ terminal of the $?_{mxl} \vars$ expression occurs at least once
inside the $YNFORM$S expression as the first argument of a $Past, Perf,
At, Before, After,$ or $Ntense$ operator, or as the second argument of a
$Part$ operator.

\item[$\gnote{3}$] : The $\vars$ expression does not occur in the $\forms$
expression.

\item[$\gnote{4}$] : If the $\terms$ expression is a $\vars$, then the
$\vars$ expression does not occur in the $\ynforms$ expression.

\end{itemize}

\subsection{Semantics of TOP}

\subsubsection*{Ontology}\label{temporal_ontology}

\paragraph{Point structure:} A \emph{point structure} for \topl is an
ordered pair $\tup{\pts, \prec}$, such that \pts 
\index{pts@\pts (set of all time-points)}
is a non-empty set, $\prec$ \index{<@$\prec$ (precedes)} is a binary
relation over $\pts \times \pts$, and $\tup{\pts, \prec}$ has the
following five properties: 
\begin{description}

\item[transitivity:] If $t_1, t_2, t_3 \in \pts$, $t_1 \prec t_2$, and
$t_2 \prec t_3$, then $t_1 \prec t_3$.

\item[irreflexivity:] If $t \in \pts$, then $t \prec t$ does not hold.

\item[linearity:] If $t_1, t_2 \in \pts$ and $t_1 \not= t_2$, then
exactly one of the following holds: $t_1 \prec t_2$ or $t_2 \prec t_1$.

\item[left and right boundedness:] There is a $t_{first} \in \pts$,
\index{tfirst@$t_{first}$ (earliest time-point)} 
such that for all $t \in \pts$, $t_{first} \preceq t$. Similarly, there is
a $t_{last} \in \pts$, 
\index{tlast@$t_{last}$ (latest time-point)}
such that for all $t \in \pts$, $t \preceq t_{last}$.

\item[discreteness:] For every $t_1, t_2 \in \pts$, with $t_1 \not=
t_2$, there is at most a finite number of $t_3 \in \pts$, such
that $t_1 \prec t_3 \prec t_2$.

\end{description}
The elements of \pts are called
\emph{time-points}. The following functions are defined on
time-points:

\paragraph{prev(t) and next(t):} 
\index{prev@$prev()$ (previous time-point)} 
\index{next@$next()$ (next time-point)} 
If $t_1 \in \pts - \{t_{last}\}$, then $next(t_1)$ denotes a $t_2 \in
\pts$, such that $t_1 \prec t_2$ and for no $t_3 \in \pts$ is it true
that $t_1 \prec t_3 \prec t_2$. Similarly, if $t_1 \in \pts -
\{t_{first}\}$, then $prev(t_1)$ denotes a $t_2 \in \pts$, such that
$t_2 \prec t_1$ and for no $t_3 \in \pts$ is it true that $t_2 \prec
t_3 \prec t_1$. Here, whenever $next(t)$ is
used, it is assumed that $t \not= t_{last}$. Similarly, whenever
$prev(t)$ is used, it is assumed that $t \not= t_{first}$.

\paragraph{Periods and instantaneous periods:}
A \emph{period} $p$ over $\tup{\pts, \prec}$ is a non-empty subset of
\pts with the following property:
\begin{description}

\item[convexity:] If $t_1, t_2 \in p$, $t_3 \in \pts$, and $t_1 \prec
t_3 \prec t_2$, then $t_3 \in p$.

\end{description}
\index{periods1@$\periods_{\tup{\pts, \prec}}$ (set of all periods over $\tup{\p
ts, \prec}$)} 
$\periods_{\tup{\pts, \prec}}$ is the set of all periods over
$\tup{\pts, \prec}$.  If $p \in \periods_{\tup{\pts, \prec}}$ and $p$
contains only one time-point, then $p$ is an \emph{instantaneous
period over $\tup{\pts, \prec}$}. $\instants_{\tup{\pts, \prec}}$ 
\index{instants1@$\instants_{\tup{\pts, \prec}}$ (set of all instantaneous perio
ds over $\tup{\pts, \prec}$)}
is the set of all instantaneous periods over $\tup{\pts, \prec}$. For
simplicity, we may write \periods 
\index{periods@$\periods$ (set of all periods)} 
and \instants  
\index{instants@$\instants$ (set of all instantaneous periods)}
instead of $\periods_{\tup{\pts, \prec}}$ and $\instants_{\tup{\pts,
\prec}}$.
\index{periods*@$\periods^*$ ($\periods \union \emptyset$)}
$\periods^*_{\tup{\pts, \prec}}$ (or simply $\periods^*$) 
is the set $\periods \union \{\emptyset\}$.

\paragraph{Subperiods:} $p_1$ is a \emph{subperiod} of
$p_2$, iff $p_1, p_2 \in \periods$ and $p_1 \subseteq p_2$. In this
case we write $p_1 \subper p_2$. 
\index{<sq@$\subper$ (subperiod)}
($p_1 \subseteq p_2$ is weaker than $p_1 \subper p_2$, because it
does not guarantee that $p_1, p_2 \in \periods$.)
Similarly, $p_1$ is a \emph{proper 
subperiod} of $p_2$, iff $p_1, p_2 \in \periods$ and $p_1 \subset
p_2$. In this case we write $p_1 \propsubper
p_2$. 
\index{<sq@$\propsubper$ (proper subperiod)}

\paragraph{Maximal periods:} 
\index{mxlpers@$mxlpers()$ (maximal periods of a set or temporal element)}
If $S$ is a set of periods, then $\mxlpers(S)$ is the set of
\emph{maximal periods} of $S$.  $\mxlpers(S) \defeq \{p \in S \mid
\text{for no } p' \in S \text{ is it true that } p \propsubper p'\}$.

\paragraph{minpt(S) and maxpt(S):} 
\index{minpt@$minpt()$ (earliest time-point in a set)}
\index{maxpt@$maxpt()$ (latest time-point in a set)}
If $S \subseteq \pts$, $minpt(S)$ denotes
the time-point $t \in S$, such that for every $t' \in S$, 
$t \preceq t'$. Similarly, if $S \subseteq \pts$, $maxpt(S)$
denotes the time-point $t \in S$, such that for every $t' \in S$, 
$t' \preceq t$. 

\paragraph{Notation:} Following standard conventions, $[t_1,
t_2]$ denotes the set $\{t \in \pts \mid t_1 \preceq t \preceq t_2
\}$. Similarly, $(t_1,
t_2]$ denotes the set $\{t \in \pts \mid t_1 \prec t \preceq t_2
\}$. $[t_1, t_2)$ and $(t_1,t_2)$ are defined similarly.

\subsubsection*{Model structure}

\label{top_model}

A \topl model $M$ is an ordered 7-tuple:
\[ M = \tup{\tup{\pts, \prec}, \objs, 
            \fcons, \fpfuns, \fculms, \fgparts, \fcparts}
\]
such that $\tup{\pts, \prec}$ is a point structure for \topl 
(section \ref{temporal_ontology}), $\periods_{\tup{\pts, \prec}}
\subseteq \objs$, and \fcons, \fpfuns, \fculms, \fgparts, and \fcparts
are as specified below.
 
\paragraph{$\mathbf{OBJS}$:} 
\index{objs@\objs (\topl's world objects)}
\objs is a set containing all the objects in the
modelled world that can be denoted by \topl terms.

\paragraph{$\mathbf{f_{cons}}$:}
\index{fcons@$\fcons()$ (maps \topl constants to world objects)}
\fcons is a function $\cons \mapsto \objs$, i.e. \fcons maps each \topl
constant to a world object.

\paragraph{$\mathbf{f_{pfuns}}$:}
\index{fpfuns@$\fpfuns()$ (returns the maximal periods where predicates hold)}
\fpfuns is a function that maps each pair $\tup{\pi, n}$, where
$\pi \in \pfuns$ and $n \in \{1,2,3,\dots\}$, to another function
$(\objs)^n \mapsto \pow(\periods)$ (where \pow(S)
denotes the powerset of S).

For every $\pi \in \pfuns$ and $n \in \{1,2,3,\dots\}$, 
$\fpfuns(\pi, n)$ must have the following property: for every
$\tup{o_1,o_2,\dots,o_n} \in (\objs)^n$, it must be
the case that: 
\[
\text{if } p_1, p_2 \in \fpfuns(\pi, n)(o_1,o_2,\dots,o_n) \text{ and }
p_1 \union p_2 \in \periods, \text{ then } p_1 = p_2
\]
Intuitively, \fpfuns shows the maximal periods where the
situation represented by $\pi(\tau_1, \dots, \tau_n)$ holds, for every
combination of arguments $\tau_1, \dots, \tau_n$.

\paragraph{$\mathbf{f_{culms}}$:}
\index{fculms@$\fculms()$ (shows if the situation of a predicate reaches its cli
max)}
\fculms is a function that maps each pair $\tup{\pi, n}$, where $\pi
\in \pfuns$ and $n \in \{1,2,3,\dots\}$, to another function $(\objs)^n
\mapsto \{T,F\}$.
Intuitively, \fculms
shows whether or not a situation reaches a climax at the latest
time-point where the situation is ongoing.

\paragraph{$\mathbf{f_{gparts}}$:}
\index{fgparts@$\fgparts()$ (assigns gappy partitionings to elements of \gparts)
}
\fgparts is a function that maps each element of \gparts to a
\emph{gappy partitioning}. A gappy partitioning is a subset $S$ of
\periods, such that for every $p_1, p_2 \in S$, $p_1 \intersect p_2 =
\emptyset$, and $\bigcup_{p \in S}p \not= \pts$.

\paragraph{$\mathbf{f_{cparts}}$:}
\index{fcparts@$\fcparts()$ (assigns complete partitionings to elements of \cpar
ts)}
\fcparts is a function that maps each element of \cparts to a
\emph{complete partitioning}. A complete partitioning is a subset
$S$ of \periods, such that for every $p_1, p_2 \in S$, $p_1 \intersect
p_2 = \emptyset$, and $\bigcup_{p \in S}p = \pts$.

\subsubsection*{Variable assignment} \label{var_assign}

A variable assignment w.r.t. a \topl model $M$ 
is a function $g: \vars \mapsto \objs$
\index{g@$g()$, $g^\beta_o()$ (variable assignment)}
($g$ assigns to each variable an element of \objs). $G_M$,
or simply $G$, 
\index{G@$G$, $G_M$ (set of all variable assignments)}
is the set of all possible variable assignments w.r.t.\ $M$.

If $g \in G$, $\beta \in \vars$, and $o \in \objs$, then $g^\beta_o$
\index{g@$g()$, $g^\beta_o()$ (variable assignment)}
is the variable assignment defined as follows: $g^\beta_o(\beta) = o$,
and for every $\beta' \in \vars$ with $\beta' \not= \beta$,
$g^\beta_o(\beta') = g(\beta)$.

\subsubsection*{Denotation of a TOP expression} \label{denotation}

In the following, we shall use the non-terminal symbols of the \topl
syntax given in section \ref{top-syntax} above as the names of sets
containing expressions which can be analysed syntactically as the 
corresponding non-terminal.
That is, where we stipulate, for example, that ``$\phi \in \ynforms$'',
this means that ``$\phi$ is a \topl expression which syntactically
is a $\ynforms$''.
Also, we will assume that in the semantic definitions for complex
expressions, all the various sub-expressions are
of the correct syntactic types, as given by the grammar in section
\ref{top-syntax} above.

An \emph{index of evaluation} is an ordered 3-tuple
$\tup{st,et,lt}$, such that $st \in \pts$, $et \in \periods$, and $lt
\in \periods^*$. 

The \emph{denotation} of a \topl expression $\xi$ w.r.t.\ a model $M$,
an index of evaluation $\tup{st,et,lt}$, and a variable assignment $g$,
is written $\denot{M,st,et,lt,g}{\xi}$ or simply
$\denot{st,et,lt,g}{\xi}$. When the denotation of $\xi$ does not
depend on $st$, $et$, and $lt$, we may write $\denot{M,g}{\xi}$
or simply $\denot{g}{\xi}$. 
\begin{itemize}

\item If $\kappa \in \cons$, then $\denot{g}{\kappa} =
  \fcons(\kappa)$.

\item If $\beta \in \vars$, then $\denot{g}{\beta} =
  g(\beta)$.

\item If $\phi \in \ynforms$, then $\denot{st,et,lt,g}{\phi} \in
\{T,F\}$. 
\end{itemize}

\begin{itemize}
\item If $\pi(\tau_1, \tau_2, \dots, \tau_n) \in \literal$,
then 
$\denot{st,et,lt,g}{\pi(\tau_1, \tau_2, \dots, \tau_n)} = T$ 
iff $et \subper lt$ and for some $p_{mxl} \in \fpfuns(\pi,
n)(\denot{g}{\tau_1}, \denot{g}{\tau_2}, \dots, \denot{g}{\tau_n})$,
$et \subper p_{mxl}$.

\end{itemize}

\begin{itemize}
\item If $\phi_1, \phi_2 \in \ynforms$, then
\index{^@$\land$ (\topl's conjunction)}
$\denot{st,et,lt,g}{\phi_1 \land \phi_2} = T$ iff 
$\denot{st,et,lt,g}{\phi_1} = T$ and $\denot{st,et,lt,g}{\phi_2} = T$. 

\item 
\index{part@$\partop[\;]$ (used to select periods from partitionings)} 
$\denot{g}{\partop[\sigma, \beta, \nu_{ord}]}$ is $T$, iff
all the following hold (below $f = \fcparts$ if $\sigma \in
\cparts$, and $f = \fgparts$ if $\sigma \in \gparts$):
  \begin{itemize}
  \item $g(\beta) \in f(\sigma)$, 

  \item if $\nu_{ord} = 0$, then $st \in g(\beta)$, 

  \item if $\nu_{ord} \leq -1$, then the
    following set contains exactly $-\nu_{ord} - 1$ elements:
    \[
    \{ p \in f(\sigma) \mid
       maxpt(g(\beta)) \prec minpt(p) \text{ and } maxpt(p) \prec st \} 
    \]
  \end{itemize}
\end{itemize}

\begin{itemize}
\index{part@$\partop[\;]$ (used to select periods from partitionings)} 
\item$\denot{g}{\partop[\sigma, \beta]} = T$ iff $g(\beta)
\in f(\sigma)$ (where $f = \fcparts$ if $\sigma \in \cparts$, and $f
= \fgparts$ if $\sigma \in \gparts$).
\end{itemize}

\begin{itemize}
\item 
\index{pres@$\pres[\;]$ (used to refer to the present)}
$\denot{st,et,lt,g}{\pres[\phi]} = T$, iff $st \in et$ and
  $\denot{st,et,lt,g}{\phi} = T$. 
\end{itemize}
\begin{itemize}
\item 
\index{past@$\past[\;]$ (used to refer to the past)}
$\denot{st,et,lt,g}{\past[\beta, \phi]} = T$, iff 
  $g(\beta) = et$ and 
  $\denot{st,et, lt \intersect [t_{first}, st), g}{\phi} = T$.
\end{itemize}
\begin{itemize}
\item 
\index{culm@$\culm[\;]$ (used to express non-progressives of culminating activit
y verbs)}
$\denot{st,et,lt,g}{\culm[\pi(\tau_1, \dots, \tau_n)]} = T$, iff
$et \subper lt$, $\fculms(\pi,n)(\denot{g}{\tau_1}, \dots,
\denot{g}{\tau_n}) = T$, $S \not= \emptyset$, and $et = [minpt(S),
maxpt(S)]$, where: 
\[
S = \bigcup_{p \in \fpfuns(\pi, n)(\denot{g}{\tau_1}, \dots,
             \denot{g}{\tau_n})}p
\]
\end{itemize}
\begin{itemize}
\item 
\index{at@$\at[\;]$ (narrows the localisation time)}
   $\denot{st,et,lt,g}{\at[\tau, \phi]} = T$, iff 
   $\denot{g}{\tau} \in \periods$ and 
   $\denot{st,et,lt \intersect \denot{g}{\tau},g}{\phi} = T$.
\item 
\index{at@$\at[\;]$ (narrows the localisation time)}
$\denot{st,et,lt,g}{\at[\phi_1, \phi_2]} = T$, iff
   for some $et'$, \\
   $et' \in mxlpers(\{e \in \periods \mid \denot{st,e,\pts,g}{\phi_1} = T\})$ 
   and 
   $\denot{st,et,lt \intersect et',g}{\phi_2} = T$.
\end{itemize}
\begin{itemize}
\item 
\index{before@$\before[\;]$ (used to express \qit{before})}
   $\denot{st,et,lt,g}{\before[\tau, \phi]} = T$, iff 
   $\denot{g}{\tau} \in \periods$ and 
   $\denot{st,et,
           lt \intersect [t_{first}, minpt(\denot{g}{\tau})),
           g}{\phi} = T$.
\item 
\index{before@$\before[\;]$ (used to express \qit{before})}
   $\denot{st,et,lt,g}{\before[\phi_1, \phi_2]} = T$, iff
   for some $et'$, \\
   $et' \in mxlpers(\{e \in \periods \mid \denot{st,e,\pts,g}{\phi_1} = T\})$,
   and
   $\denot{st,et,
           lt \intersect [t_{first}, minpt(et')),
           g}{\phi_2} = T$.
\item 
\index{after@$\after[\;]$ (used to express \qit{after})}
   $\denot{st,et,lt,g}{\after[\tau, \phi]} = T$, iff 
   $\denot{g}{\tau} \in \periods$ and 
   $\denot{st,et,
           lt \intersect (maxpt(\denot{g}{\tau}), t_{last}],
           g}{\phi} = T$.
\item 
\index{after@$\after[\;]$ (used to express \qit{after})}
   $\denot{st,et,lt,g}{\after[\phi_1, \phi_2]} = T$, iff
   for some $et'$,
   $et' \in mxlpers(\{e \mid \denot{st,e,\pts,g}{\phi_1} = T\})$
   and
   $\denot{st,et,
           lt \intersect (maxpt(et'), t_{last}],
           g}{\phi_2} = T$.
\end{itemize}
\begin{itemize}
\item 
\index{fills@$\fills[\;]$ (requires $et = lt$)}
   $\denot{st,et,lt,g}{\fills[\phi]} = T$, iff $et = lt$ and
   $\denot{st,et,lt,g}{\phi} = T$.  
\end{itemize}
\begin{itemize}
\item 
\index{begin@$\lbegin[\;]$ (used to refer to start-points of situations)}
   $\denot{st,et,lt,g}{\lbegin[\phi]} = T$, iff $et \subper lt$ and
for some $et'$\\
   $et' \in mxlpers(\{e \in \periods \mid \denot{st,e,\pts,g}{\phi} = T\})$ 
   and $et = \{minpt(et')\}$.
\item 
\index{end@$\lend[\;]$ (used to refer to end-points of situations)}
   $\denot{st,et,lt,g}{\lend[\phi]} = T$, iff $et \subper lt$ and 
for some $et'$\\
   $et' \in mxlpers(\{e \in \periods \mid \denot{st,e,\pts,g}{\phi} = T\})$ 
   and $et = \{maxpt(et')\}$.
\end{itemize}
\begin{itemize}
\item 
\index{ntense@$\ntense[\;]$ (used when expressing nouns or adjectives)}
  $\denot{st,et,lt,g}{\ntense[\beta, \phi]} = T$, iff for some
  $et' \in \periods$, $g(\beta)= et'$ and
  $\denot{st,et',\pts,g}{\phi} = T$. 
\item 
\index{ntense@$\ntense[\;]$ (used when expressing nouns or adjectives)}
  $\denot{st,et,lt,g}{\ntense[now^*, \phi]} = T$, iff 
  $\denot{st,\{st\},\pts,g}{\phi} = T$.
\end{itemize}

\begin{itemize}
\item 
\index{for@$\for[\;]$ (used to express durations)}
$\denot{st,et,lt,g}{\for[\sigma_c, \nu_{qty}, \phi]} = T$, iff
$\denot{st,et,lt,g}{\phi} = T$, and for some $p_1,p_2,\dots,p_{\nu_{qty}}
\in \fcparts(\sigma_c)$, it is true that $minpt(p_1) = minpt(et)$,
$next(maxpt(p_1)) = minpt(p_2)$, $next(maxpt(p_2)) = minpt(p_3)$,
\dots, $next(maxpt(p_{\nu_{qty} - 1})) = minpt(p_{\nu_{qty}})$, and
$maxpt(p_{\nu_{qty}}) = maxpt(et)$.
\end{itemize}

\begin{itemize}
\item 
\index{perf@$\perf[\;]$ (used to express the past perfect)}
$\denot{st,et,lt,g}{\perf[\beta, \phi]} = T$, iff $et \subper
lt$, and for some $et' \in \periods$, it is true that $g(\beta) =
et'$, $maxpt(et') \prec minpt(et)$, and $\denot{st,et',\pts,g}{\phi}
= T$. 
\end{itemize}

\begin{itemize}
\item
 \index{?@$?$ (\topl's interrogative quantifier)}
     $\denot{st,et,lt,g}{?\beta_1 \; ?\beta_2 \; \dots \; ?\beta_n \; \phi} 
      = \{\tup{g(\beta_1), g(\beta_2), \dots, g(\beta_n)} \mid
          \denot{st,et,lt,g}{\phi} = T\}$
\end{itemize}
\begin{itemize}
\item 
\index{?@$?$ (\topl's interrogative quantifier)}
\index{?mxl@$?_{mxl}$ (\topl's interrogative-maximal quantifier)}
      $
      \denot{st,et,lt,g}{?_{mxl}\beta_1 \; ?\beta_2 \; ?\beta_3 \;
      \dots \; ?\beta_n \; \phi} = 
      $ 
      $
      \{\tup{g(\beta_1), g(\beta_2), g(\beta_3), \dots, g(\beta_n)} \mid 
      \denot{st,et,lt,g}{\phi} = T,
      \text{and }
      $
      $
      \text{ for no } et' \in \periods \text{ and } g' \in G 
      \text{ is it true that  } 
      \denot{st,et',lt,g'}{\phi} = T, 
      $ 
      $\ 
      g(\beta_1) \propsubper g'(\beta_1), \; 
      g(\beta_2) = g'(\beta_2), \; g(\beta_3) = g'(\beta_3), \;
      \dots, \; g(\beta_n) = g'(\beta_n)\}
      $
\end{itemize}

\par\noindent
The denotation of $\phi$ w.r.t.\ $M, st$, written $\denot{M,st}{\phi}$
or simply $\denot{st}{\phi}$, is defined only for \topl formulae:
\begin{itemize}

\item If $\phi \in \ynforms$, then $\denot{st}{\phi} =$
\begin{itemize}
\item $T$, if for some $g \in G$ and $et \in \periods$, 
$\denot{st,et,\pts,g}{\phi} = T$, 
\item $F$, otherwise
\end{itemize}

\item If $\phi \in \whforms$, then $\denot{st}{\phi} = 
   \bigcup_{g \in G, \; et \in \periods}\denot{st,et,\pts,g}{\phi}$. 
\end{itemize}
Within the \nltdb, 
each yes/no question (e.g. \qit{Is tank 2 empty?})
is mapped to a $\phi \in \ynforms$ (multiple formulae are generated for
ambiguous questions).  The answer is affirmative iff 
$\denot{st}{\phi} = T$. 
With questions containing wh-words (e.g. \qit{Who repaired what?}), the
question is mapped to a $\phi \in \whforms$ (again multiple formulae are
generated if the question is ambiguous), and the answer reports the
tuples in $\denot{st}{\phi}$.

\end{document}